%% file: main.tex
\documentclass[11pt, letterpaper]{cmu}

\usepackage[all]{hypcap}
\usepackage[comma,numbers,sort,compress]{natbib}
\usepackage{hyperref}[citecolor=magenta]

\hypersetup{
    colorlinks = true,
    citecolor = {magenta},
}

\usepackage{microtype}
\usepackage{graphicx}
\usepackage{subfigure}
\usepackage{booktabs}
\usepackage{float}
\usepackage{amsmath}
\usepackage{amssymb}
\usepackage{mathtools}
\usepackage{amsthm}
\usepackage{mathrsfs}
\usepackage{nicefrac}
\usepackage{dsfont}
\usepackage{enumitem}
\usepackage{float}
\usepackage[utf8]{inputenc}

\usepackage{threeparttable}
\usepackage{booktabs}

\definecolor{myblue}{rgb}{0.1,0.4,0.9}

  {\end{tcolorbox}}
\usepackage[capitalize,noabbrev]{cleveref}
\usepackage{subcaption}
\usepackage{wrapfig}
\usepackage{lipsum}
\usepackage{listings}

\usepackage{amsmath}
\usepackage{amssymb}
\usepackage{mathtools}
\usepackage{amsthm}
\usepackage{bbm}
\usepackage{fleqn, tabularx}
\usepackage{multirow}
\usepackage[export]{adjustbox}
\usepackage{colortbl}

\usepackage{algpseudocode}
\usepackage{setspace}

\usepackage{color}
\definecolor{deepblue}{rgb}{0,0,0.5}
\definecolor{deepred}{rgb}{0.6,0,0}
\definecolor{deepgreen}{rgb}{0,0.5,0}
\definecolor{boost_correct_to_correct}{HTML}{66C2A5}
\definecolor{default_correct_to_correct}{HTML}{fc8d62}
\definecolor{dup_correct_to_correct}{HTML}{8da0cb}
\definecolor{new_correct_to_correct}{HTML}{e78ac3}

\newcommand\pythonstyle{\lstset{
basicstyle=\ttfamily\footnotesize,
language=Python,
morekeywords={self, clip, exp, mse_loss, uniform_sample, concatenate, logsumexp},              
keywordstyle=\color{deepblue},
emph={MyClass,__init__},          
emphstyle=\color{deepred},   
stringstyle=\color{deepgreen},
frame=single,                       
showstringspaces=false
}}

\lstnewenvironment{python}[1][]
{
\pythonstyle
\lstset{#1}
}
{}

\newcommand\pythoninline[1]{{\pythonstyle\lstinline!#1!}}

\input{math_commands}

\makeatletter
\def\mathcolor#1#{\@mathcolor{#1}}
\def\@mathcolor#1#2#3{%
  \protect\leavevmode
  \begingroup
    \color#1{#2}#3%
  \endgroup
}
\makeatother

\usepackage[textsize=tiny]{todonotes}
\usepackage{algorithm}
\usepackage{amssymb}

\usepackage[skins,theorems]{tcolorbox}

\usepackage{crossreftools}
\usepackage[suppress]{color-edits}

\usepackage{dsfont}
\usepackage{nicefrac}

\newtcolorbox{analysisbox}[1][]{
    enhanced jigsaw,
    colback=white,
    colframe=blue!75!black,
    fonttitle=\bfseries,
    boxsep=5pt,
    left=5pt,
    right=5pt,
    top=5pt,
    bottom=5pt,
    title=#1,
}

\tcbset{
  aibox/.style={
    width=\linewidth,
    top=8pt,
    bottom=4pt,
    colback=rliableolive!8!white,
    colframe=black,
    colbacktitle=black,
    enhanced,
    center,
    attach boxed title to top left={yshift=-0.1in,xshift=0.15in},
    boxed title style={boxrule=0pt,colframe=white,},
  }
}
\newtcolorbox{AIbox}[2][]{aibox,title=#2,#1}
\definecolor{lightblue}{rgb}{0.22,0.45,0.70}% light blue

\newcommand{\nauman}[1]{ \textcolor{blue}{\small [MN: #1]}}

\usepackage{algorithm}      % float wrapper for algorithms
\usepackage{algpseudocode}  % pseudocode commands (\For, \Function, etc.)
\usepackage[endLComment=,italicComments=false]{algpseudocodex} % nice extras
\usepackage{tcolorbox}      % boxes
\usepackage{xcolor}
\usepackage{pifont}
\usepackage{soul}
\definecolor{highlightmistake}{RGB}{255, 179, 179}
\definecolor{highlightcorrect}{RGB}{179, 255, 179} 

\definecolor{rliableolive}{HTML}{BBCC33}
\definecolor{rliableblue}{HTML}{77AADD}
\definecolor{rliablered}{HTML}{EE8866}

\newenvironment{olivebox}{%
    \begin{tcolorbox}[colback=rliableolive!10!white,colframe=black,boxsep=3pt,top=4pt,bottom=4pt,left=3pt,right=3pt]
}{%
    \end{tcolorbox}
}
\definecolor{myblue}{rgb}{0.1,0.4,0.9}

\title{\methodname{}: Training Critics via Flow-Matching for Scaling Compute in Value-Based RL}

\author[1]{Bhavya Agrawalla}
\author[2]{Michal Nauman}
\author[1]{Khush Agrawal}
\author[1]{Aviral Kumar}

\affil[1]{Carnegie Mellon University}
\affil[2]{University of Warsaw}

\correspondingauthor{bbagrawa@andrew.cmu.edu}

\begin{abstract}
\textbf{Abstract:} A hallmark of modern large-scale machine learning techniques is the use of training objectives that provide dense supervision to intermediate computations, such as teacher forcing the next token in language models or denoising  step-by-step in diffusion models. This enables models to learn complex functions in a generalizable manner. Motivated by this observation, we investigate the benefits of iterative computation for temporal difference (TD) methods in reinforcement learning (RL). Typically they represent value functions in a monolithic fashion, without iterative compute. We introduce \methodname{} (\emph{flow-matching Q-functions}), an approach that parameterizes the Q-function using a velocity field and trains it using techniques from flow-matching, typically used in generative modeling. This velocity field underneath the flow is trained using a TD-learning objective, which bootstraps from values produced by a target velocity field, computed by running multiple steps of numerical integration. Crucially, \methodname{} allows for more fine-grained control and scaling of the Q-function capacity than monolithic architectures, by appropriately setting the number of integration steps. Across a suite of challenging offline RL benchmarks and online fine-tuning tasks, \methodname{} improves performance by nearly 1.8$\times$. \methodname{} scales capacity far better than standard TD-learning architectures, highlighting the potential of iterative computation for value learning.
\\

\textbf{Code and runs for \methodname{} can be found at}: \href{https://github.com/CMU-AIRe/floq}{https://github.com/CMU-AIRe/floq}
\end{abstract}

\begin{document}

\maketitle

\vspace{-0.3cm}
\section{Introduction}
\label{sec:introduction}
\vspace{-0.2cm}

\begin{wrapfigure}{r}{0.5\textwidth}
\vspace{-0.8cm}
\centering
\includegraphics[width=0.99\linewidth]{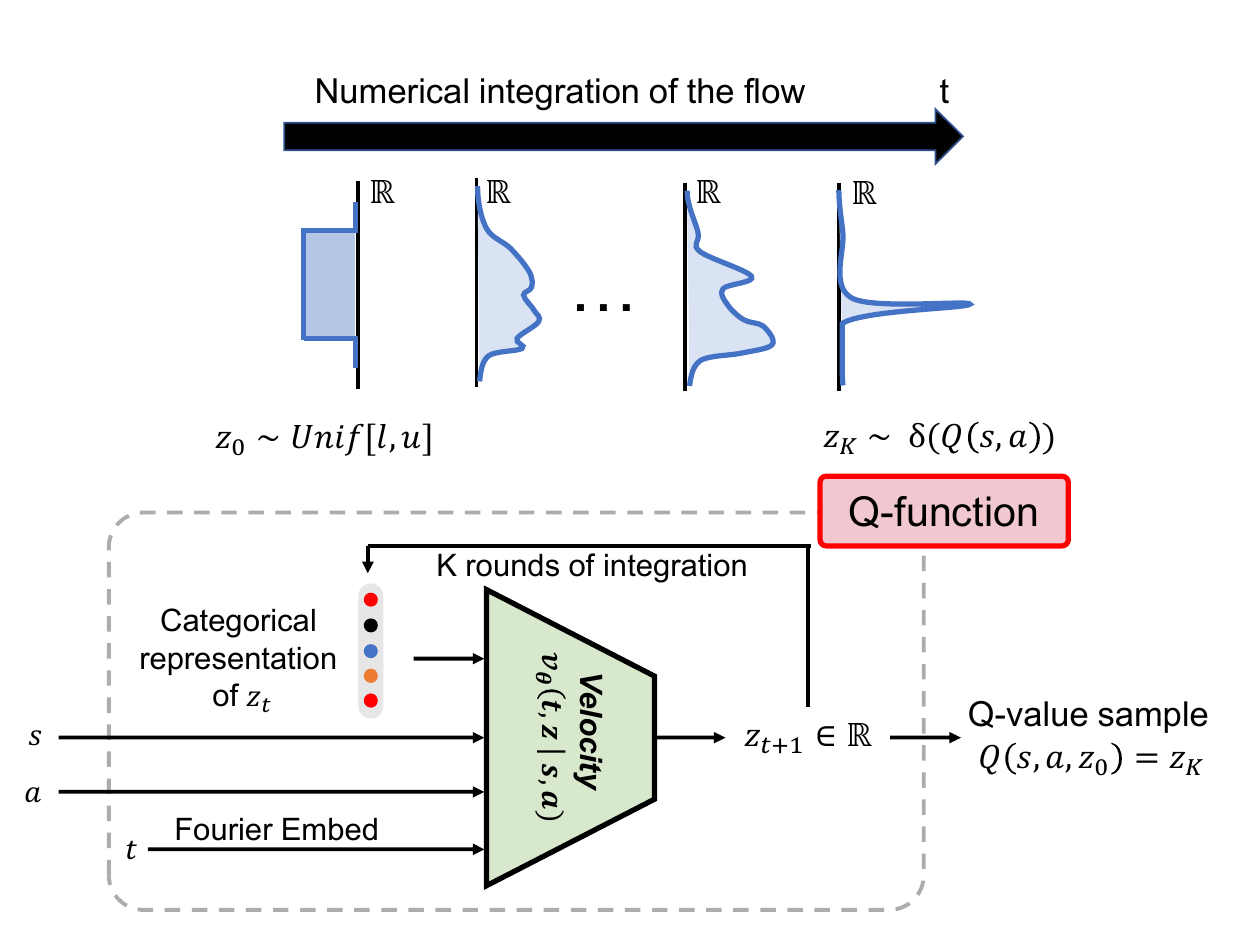}
\vspace{-0.65cm}
    \caption{\footnotesize{\textbf{\methodname{} architecture.} Our approach models the Q-function via a velocity field in a flow-matching generative model. Over multiple calls, this velocity field converts a randomly sampled input $\bz(0)$ into a sample from the Dirac-delta distribution centered at the Q-value. We prescribe how this sample can be trained via a flow-matching loss. Doing this enables us to scale computation by running numerical integration, with multiple calls to the velocity field. To train \methodname{}, we utilize a categorical representation of input $\bz_t$~\citep{farebrother2024stop}.}}    \label{fig:floq_main}
    \vspace{-0.7cm}
\end{wrapfigure}
A key principle that has enabled effective model scaling in various areas of machine learning is the use of \textbf{\emph{iterative computation}}: producing complex output functions by composing a sequence of small and simple operations. For example, transformers~\citep{vaswani2017attention} are able to generate coherent text and long reasoning chains by predicting the next token and by composing multiple atomic reasoning strategies~\citep{gandhi2025cognitivebehaviorsenableselfimproving} respectively. Similarly, diffusion models~\citep{ddpm_ho2020,diffusion_sohl2015} and flow-matching techniques~\citep{flow_lipman2023,flow_albergo2023} learn to synthesize realistic images by progressively denoising small perturbations. The wide application of these models suggests that iterative computation is a powerful tool for modeling complex functions with deep networks.

Motivated by these empirical results, in this paper, we ask: \textbf{\emph{can iterative computation also improve value estimation in reinforcement learning (RL)?}} Specifically, we are interested in improving the estimation of the Q-value function. While Q-functions map state-action inputs to a scalar value, they are known to be highly complex and difficult to fit accurately (e.g.,~\citep{dong2020expressivity}). Standard temporal-difference (TD) learning used to train Q-functions struggles to leverage capacity of deep networks~\citep{kumar2021implicit,kumar2021dr3, bjorck2021towards, lyle2022learning,gulcehre2022empirical}, often resulting in poor generalization. These problems are further exacerbated in the offline RL problem setting~\citep{levine2020offline,kumar2019stabilizing}, where we must learn entirely from static datasets. This motivates exploring architectures that spend compute iteratively to estimate value functions, potentially yielding more accurate Q-values, and better downstream policies.

A natural starting point to use iterative computation for value-based RL is to represent the Q-function with ResNets~\citep{resnet}, where stacking residual blocks iteratively computes the Q-value. Recent work in TD-learning has obtained modest gains with ResNets~\citep{kumar2023offline,kumar2022pre,farebrother2024stop,nauman2024bigger}, but these methods deliberately need to operate with normalization and regularizers to enable stable training~\citep{bjorck2021towards, nauman2024bigger,lee2024simba,lee2025hyperspherical,kumar2023offline}. Despite improvements, these approaches lack one ingredient that makes iterative computation effective in transformers or diffusion models: \textbf{\emph{supervision at every step}} of the iterative computation process. Just as next-token prediction supervises each generated token, and diffusion supervises each denoising step, we hypothesize that stepwise loss supervision applied to TD learning might lead to improvements. 

With this observation, to effectively leverage iterative computation with dense supervision, we design a novel architecture for representing Q-functions. Instead of using a single monolithic network, we represent the Q-function as a \emph{velocity field} over a scalar value (Figure~\ref{fig:floq_main}). Our approach, \methodname{} (flow-matching Q-functions) samples a scalar uniformly distributed noise variable and maps it to the Q-value by numerically integrating the predictions of this velocity field. We train this velocity field with a linear flow-matching objective~\citep{flow_albergo2023,flow_lipman2023} adapted from generative modeling. Unlike standard flow-matching, which supervises the velocity with stationary targets, we train the velocity to match the evolving TD-targets. At each step, we minimize the deviation between the current Q-value estimate and the corresponding TD-target. We introduce several design choices that stabilize training and help the architecture scale capacity effectively. First, we appropriately set the support of the initial noise to be as wide as possible. Second, we use a categorical input representation to handle the non-stationarity of inputs during training. Finally, we apply a tailored Fourier-basis embedding of time so that the velocity predictions vary meaningfully across integration steps. These design choices are crucial for learning a good \methodname{} critic.

We use \methodname{} to represent the Q-function for a number of complex RL~\citep{levine2020offline,kumar2019stabilizing} tasks from the OGBench~\citep{ogbench_park2025} benchmark, previously studied by \citet{park2025flow}. In aggregate, we find that \methodname{} outperforms offline RL algorithms that represent Q-functions using a monolithic network by nearly $1.8\times$. \methodname{} is superior even when these approaches are provided with more parameters, and more complex and higher capacity architectures. \methodname{} also outperforms existing methods when running online fine-tuning after offline RL pre-training. We also show that increasing the number of flow-matching steps results in better downstream policy performance. Allocating the same capacity via Q-network ensembles or ResNets performs worse, even in a compute-matched evaluation. Finally, our approach enables a pathway for ``test-time scaling''~\citep{snell2024scaling} by increasing the number of integration steps to calculate Q functions.

\vspace{-0.2cm}
\section{Related Work}
\label{sec:related}
\vspace{-0.2cm}

\textbf{Expressive generative policies in RL.}
A growing body of work in offline RL has emphasized the importance of moving beyond unimodal Gaussian policies~\citep{kumar2019stabilizing,fujimoto2018off,kumar2020conservative,fujimoto2021minimalist} toward richer, generative function classes such as diffusion models~\citep{wang2022diffusion,hansen2023idql,yang2023policy, bansal2023cold, li2024learning,ren2024diffusion}, flow-based policies~\citep{flow_lipman2023, flow_albergo2023, park2025flow}, and sequence models~\citep{janner2021sequence,lee2022multi,yamagata2023q}. This shift is motivated by evidence that policy learning is often a significant bottleneck in offline RL~\citep{kostrikov2021iql, park2024value}. Diffusion policies have been among the first expressive alternatives, but early variants such as Diffusion-QL~\citep{wang2022diffusion} faced optimization challenges. Subsequent work has explored more stable formulations through implicit action re-ranking~\citep{hansen2023idql}, action-space gradient estimators~\citep{yang2023policy,li2024learning}, or diffusion policy gradient methods~\citep{ren2024diffusion}. Flow-matching policies have recently emerged as another promising direction~\citep{park2025flow}. In parallel, policy-agnostic frameworks such as PA-RL~\citep{parl_mark2024} have sought to decouple algorithmic progress from specific architectural choices, enabling the use of diffusion, flows, or transformers interchangeably. Finally, transformer-based action models~\citep{janner2021sequence,lee2022multi,yamagata2023q} have proven effective in imitation and sequence modeling settings, though their integration with RL objectives remains less mature~\citep{kumar2023offline}. Complementarily, we do not focus exclusively on policy expressivity and instead aim to utilize more expressive Q-functions, while preserving compatibility with diverse policy learning strategies.

\textbf{Scaling Q-functions.} Efforts to scale Q-functions in RL have taken multiple directions with novel training objectives, architectures, and regularization strategies. One line of work reformulates TD-learning by replacing mean-squared error regression with classification-based objectives~\citep{kumar2023offline,farebrother2024stop, nauman2025bigger, seo2025fasttd3}, which have been shown to improve stability in multi-task problems. These approaches, however, typically retain standard backbones such as ResNets~\citep{resnet,kumar2022pre} or Transformers~\citep{vaswani2017attention}. A second line of research emphasizes regularization~\citep{kumar2021implicit, lyle2021effect, kumar2021dr3}, to mitigate overfitting and other pathologies arising when training large models with TD learning. Beyond these, several studies propose explicit architectural modifications, such as normalization layers~\citep{nauman2024bigger}, action discretization schemes~\citep{qt_chebotar2023}, perceive-actor-critic hybrids~\citep{pac_springenberg2024}, or mixture-of-experts formulations~\citep{obando2024mixtures}. Previous work has also attempted to develop scaling laws for TD learning~\citep{rybkin2025value, fu2025compute} and showing that alternatives to TD, such as contrastive RL~\citep{crl_eysenbach2022}, can scale to deeper architectures~\citep{wang20251000}. Despite these advances, a clear recipe for scaling value-based RL with TD-learning has yet to emerge. In this context, our work demonstrates that compute-efficient scaling can be realized not simply by increasing depth or width, but by introducing dense intermediate supervision through multiple integration steps of the Q-function. As such, \methodname{} introduces a novel axis of scaling, allowing for compute scaling through additional integration steps rather than network depth or width.

\textbf{Test-time scaling in RL.} A complementary line of work studies how decision-time compute can be traded for performance by planning with learned world models at deployment. Classical MPC-style planners coupled with learned dynamics models such as PETS~\citep{chua2018deep}, MPPI~\citep{williams2017model}, and PDDM~\citep{nagabandi2018learning} naturally allow scaling via more samples, longer horizons, or more optimizer iterations at test time. In offline RL, MBOP~\citep{argenson2020model} explicitly adopts decision-time planning with a learned model, a behavior prior, and a terminal value to extend the effective horizon. Generative world models enable similar test-time scaling by planning inside the learned model~\citep{janner2021sequence,janner2022diffuser}. Latent-dynamics MPC approaches~\citep{tdmpc_hansen2022, tdmpc2_hansen2023} combine short-horizon optimization in latent space with a learned value for long-horizon estimates, yielding a planner that can use a variable amount of compute upon deployment. Similarly, performance of MCTS-style methods improves with more simulation~\citep{schrittwieser2020mastering,hubert2021learning, danihelka2022policy, ye2021mastering}. Across all methods listed in this paragraph, the general pattern is that increasing test-time budget (i.e. simulations, horizon, candidate trajectories) improves returns up to the limit set by model bias and value estimation error. However, none of these works use more test-time compute to better estimate a value function. Our results show that \methodname{} can not only use more integration steps at inference time to amplify the ``capacity'' of the Q-function, but also that doing so during training helps us learn better Q-functions in the first place. We believe that we are the first to show that using more integration steps is a viable and effective path to test-time compute scaling for critic networks by relying on principles of iterative computation.

\vspace{-0.2cm}
\section{Background and Preliminaries}
\label{sec:prelims}
\vspace{-0.2cm}

The goal in RL is to learn the optimal policy for an MDP $\mathcal{M} = (\mathcal{S}, \mathcal{A}, P, r, \rho, \gamma)$. $\mathcal{S}, \mathcal{A}$ denote the state and action spaces.  $P(s' | s, a)$ and $r(s,a)$ are the dynamics and reward functions. $\rho(s)$ denotes the initial state distribution.  $\gamma \in (0,1)$ denotes the discount factor. Formally, the goal is to learn a policy $\pi:\mc S\mapsto \mc A$ that maximizes cumulative discounted value function, denoted by $V^\pi(s) = {\frac{1}{1-\gamma}\sum_{t} \bb E_{a_t \sim \pi(s_t)}\brac{\gamma^t r(s_t, a_t)|s_0=s}}$. 

The Q-function of a policy $\pi$ is defined as ${Q^\pi(s,a) = {\frac{1}{1-\gamma}\sum_{t} \bb E_{a_t \sim \pi(s_t)}\brac{\gamma^t r(s_t, a_t)|s_0=s,a_0=a}}}$, and we use $Q_\theta^\pi$ to denote the estimate of the Q-function of a policy $\pi$ as obtained via a neural net with parameters $\theta$. Value-based RL methods train a Q-network by minimizing the temporal difference~(TD) error:
{
\setlength{\abovedisplayskip}{6pt}
\setlength{\belowdisplayskip}{6pt}
\begin{align}
    \!\!\!L(\theta) = \mathbb{E}_{(\bs, \ba, \bs') \sim \mathcal{D}, \ba' \sim \pi(\cdot|\bs')}\left[ \left(r(\bs, \ba) + \gamma \bar{Q}(\bs', \ba') - Q_\theta(\bs, \ba) \right)^2\right], \label{eq:td-err}
\end{align}
}where $\mathcal{D}$ is the replay buffer or an offline dataset, $\bar{Q}$ is the target Q-network, $\bs$ denotes a state, and $\ba'$ is an action drawn from a policy $\pi(\cdot | \bs)$ that aims to maximize $Q_\theta(\bs, \ba)$.
We operate in the offline RL~\citep{levine2020offline} problem setting, where the replay buffer $\mathcal{P}$ corresponds to a static dataset of transitions $\mathcal{D} = \{(s, a, r, s^\prime)\}$ collected using a behavior policy $\behavior$. Our goal in this setting is to train a good policy using the offline dataset $\mathcal{D}$ alone. The Q-network, $Q_\theta$ is typically parameterized by a deep network (e.g., an MLP).

\textbf{Offline RL algorithms.} Offline RL methods aim to learn a policy that maximizes reward while penalizing deviation from the behavior policy $\behavior$, in order to mitigate the challenge of distributional shift. This objective has been instantiated in various ways, including behavioral regularization~\citep{brac_wu2019, fujimoto2021minimalist, rebrac_tarasov2023,park2025flow, kumar2019stabilizing}, pessimistic value function regularization~\citep{kumar2020conservative}, implicit policy constraints~\citep{rwr_peters2007, peng2019advantage, crr_wang2020, parl_mark2024}, and in-sample maximization~\citep{kostrikov2021iql, sql_xu2023, xql_garg2023}. While our proposed \methodname{} architecture for Q-function parameterization is agnostic to the choice of offline RL algorithm, we instantiate it on top of FQL~\citep{park2025flow} that utilizes a flow-matching policy to better model multimodal action distributions. FQL trains the Q-function using the standard temporal-difference (TD) error from soft actor-critic (SAC)~\citep{haarnoja2018sacapps}, shown in Equation~\ref{eq:td-err}, 
% \nauman{super nitpick is that 3.1 represents regular value update, not soft value update (lacks logprob)}, 
and optimizes the policy to stay close to a behavior policy estimated via flow-matching. 
% We refer readers to \citet{park2025flow} for full details.

\textbf{Flow-matching.} Flow-matching~\citep{flow_lipman2023,liu2023flow,flow_albergo2023} is an approach for training generative models by integrating deterministic ordinary differential equations (ODEs), which also makes use of iterative computation. Flow-matching is often posed as a deterministic alternative to denoising diffusion models~\citep{ddpm_ho2020,score_song2021} that utilize stochastic differential equations (SDEs) for generating complex outputs. Concretely, given a target data distribution $p(\bx)$ over $\bx \in \mathbb{R}^d$, flow-matching attempts to fit a time-dependent \textbf{\emph{velocity field}}, $v_\theta(t, \bx) : [0,1] \times \mathbb{R}^d \to \mathbb{R}^d$ such that the solution $\psi_\theta(t, \bx)$ to the ODE:
\begin{align}
\label{eq:ode_to_solve}
\frac{d}{dt} \psi_\theta(t, \bx) = v_\theta(t, \psi_\theta(t, \bx)), \quad \psi_\theta(0, \bx(0)) = \bx(0)
\end{align}
transforms samples $\bx(0)$ from a simple base distribution (e.g., standard Gaussian or uniform, as we consider in this work) into samples from $p(\bx)$ at time $t = 1$. While there are several methods to train a velocity flow, perhaps the simplest and most widely-utilized approach is linear flow matching~\citep{flow_lipman2023}, which trains the velocity flow to predict the gradient obtained along the linear interpolating path between $\bx(0)$ and $\bx(1)$ at all intermediate points. Concretely, define $\bx(0) \sim p_0(\bx)$ be a sample from a simple initial distribution, $\bx(1) \sim p(\bx)$ be a sample from the target distribution, and $t \sim \mathrm{Unif}([0,1])$, we define interpolated points as $\bx(t) = (1 - t) \cdot \bx(0) + t \cdot \bx(1)$, and train the velocity field to minimize the squared error from the slope of the straight line connecting $\bx(0)$ and $\bx(1)$ as follows:
\begin{align}
    \label{eq:standard_flow_matching_loss}
    \min_\theta \; \mathbb{E}_{\bx(0), \bx(1), t} \left[ \left\| v_\theta(t, \bx(t)) - \frac{(\bx(1) - \bx(0))}{1 - 0} \right\|_2^2 \right].
\end{align}
After training the velocity field $v_\theta(t, \bx(t))$, flow-matching runs numerical integration to compute $\psi_\theta(t, \bx(0))$. This numerical integration procedure makes several calls to compute the velocity field. Each subsequent call runs the velocity field, $v_\theta(t, \bx{(t)})$ on input $\bx(t)$ generated as output from the previous call, representing a form of an iterative computation process.

\vspace{-0.2cm}
\section{\methodname{}: Training Q-Functions with Flow-Matching}
\vspace{-0.2cm}
In this section, we introduce the our proposed approach, \methodname{} (flow-matching Q-functions), which leverages iterative computation to train Q-functions. \methodname{} frames Q-value prediction as the task of transforming noise into a scalar Q-value. The key insight behind \methodname{} is to represent the Q-function as a neural velocity field trained via a flow matching objective. 
By parameterizing value estimation as a numerical integration process of the velocity's predictions, \methodname{} not only leverages iterative computation but this parameterization also allows us to supervise this process at all intermediate steps. We now formalize the conceptual idea behind our approach and address the two central design questions that are needed to make \methodname{} work: \textbf{(a)} how to handle moving target values in the training loss for a flow-based Q-function and \textbf{(b)} how to do effective flow-matching over scalar Q-values without collapse for learning.

\vspace{-0.25cm}
\subsection{{\methodname{} Parameterization}} 
\vspace{-0.15cm}
In contrast to standard Q-functions parameterized by deep neural networks that map state-action pairs to scalar values, \methodname{} parameterizes a time-dependent, state-action-conditioned velocity field \( v_\theta(t, \bz \mid s, a) \) over a one-dimensional latent input \( \bz \in \mathbb{R} \). At \( t = 0 \), this input \( \bz \) is sampled from the uniform distribution \( \mathrm{Unif}~[l, u] \), where $l$ and $u$ are scalars that define the range of initial sample noise used for training. The velocity field transforms the initial sample \( \bz \) into a distribution over the Q-value. We will train \methodname{} such that the learned distribution of Q-values match a Dirac-Delta around the groundtruth Q-function, i.e., $\psi_\theta(1, \bz | s, a) \sim \delta_{Q^\pi(s, a)}$ at \( t = 1 \). We can obtain the Q-value sample by numerically integrating the ODE using the Euler method. One instantiation is shown below. $\forall j \leq K$:
\begin{align}
\label{eq:compute_integral}
\psi_\theta(\nicefrac{j}{K}, \bz \mid s, a) = \bz + \frac{1}{K} \sum_{i=1}^{j} v_\theta\left(\frac{i}{K}, \psi_\theta\left(\nicefrac{i-1}{K}, \bz \mid s, a\right) \Big| s, a\right).~~ Q(s, a, \bz) := \psi_\theta(1, \bz \mid s, a)
\end{align}
An example illustration of this process is shown in Figure~\ref{fig:floq_main}. This iterative process enables us to dynamically adjust the complexity of the Q-function by varying the number of integration steps \( K \), effectively controlling the number of  evaluations of the velocity field \( v_\theta \), and thereby the  ``depth'' of the model. Finally, we remark that although Equation~\ref{eq:compute_integral} may appear similar to performing averaging like an ensemble, it is fundamentally different: the inputs passed to the velocity field \( v_\theta \) at each step $i$ depend on its own outputs from the previous step $i-1$. This recursive dependence introduces a form of iterative computation that is absent in conventional ensembles, that perform computation in parallel. As we demonstrate in our experiments (Section~\ref{sec:exps}), this formulation enjoys greater expressivity than simply ensembling multiple independently trained neural networks without iterative computation. 

In practice, the velocity field $v_\theta(\nicefrac{i}{K}, \cdot| s, a)$ can be conditioned on the various representations of the intermediate Q-values $\psi_\theta(\nicefrac{i-1}{K}, \cdot | s, a)$ to improve the effectiveness of learning. We opt to use a categorical representation of $\psi_\theta$ when passing it as input to the velocity network. We discuss this in Section~\ref{sec:practical}.

\vspace{-0.25cm}
\subsection{\textbf{Training Loss for the \methodname{} Architecture}}
\label{sec:training}
\vspace{-0.15cm}
With this parameterization in place, the next step is to design a training loss for the velocity field. Analogous to temporal-difference (TD) learning methods, a natural starting point is to iteratively train the velocity field using a loss that resembles linear flow-matching (Equation~\ref{eq:standard_flow_matching_loss}), but with targets obtained via Bellman bootstrapping. This is akin to TD-flows~\citep{farebrother2025temporal} and $\gamma$-models~\citep{janner2020gamma} that train a generative model with TD-bootstrapped targets. To do so, we introduce a target velocity field \( \tilde{v}_\theta(t, \bz \mid s, a) \), parameterized as a stale moving average of the main velocity field \( v_\theta \), similar to target networks in standard value-based RL. Given a transition \( (s, a, r, s') \), we first sample an action \( a' \sim \pi(\cdot \mid s') \) from the current policy at the next state s', and compute the target Q-value sample $\psi_{\tilde{\theta}}(1, \bz' \mid s', a')$ 
by integrating the target flow, starting from some $\bz'$ (via Euler integration) to obtain the predicted Q-value sample, $\psi_{\tilde{\theta}}(1, \bz' \mid s', a')$.
% \begin{align}
% \label{eq:compute_flow_integral}
% \psi_{\tilde{\theta}}(1, \bz' \mid s', a') = \bz' + \int_0^1 \tilde{v}_\theta(t, \psi_{\tilde{\theta}}(t, \bz' \mid s', a') \mid s', a') \, dt.
% \end{align}

We then average the predicted Q-value samples $\psi_{\tilde{\theta}}(1, \bz' \mid s', a')$ for several values of the initial noise $\bz'$ to compute an estimate of the target Q-value $Q_{\tilde{\theta}}(s', a')$. 
The bootstrapped TD-target is given by: \( y(s, a) = r(s, a) + \gamma \frac{1}{m} \sum_{j=1}^m \psi_{\tilde{\theta}}(1, \bz'_j \mid s', a') \), where $r(s, a)$ denotes the reward estimate for transition. 

We are now able to train the Q-value at state-action pair $(s, a)$ by regressing to the target value $y(s, a)$ using flow matching. Concretely, given a \( t \sim \mathrm{Unif}[0, 1] \), we construct an \textbf{interpolant} between noise $\bz$ sampled at the initial step and the target Q-value $y$, \( \bz(t) = (1 - t) \cdot \bz + t \cdot {y}(s, a) \), and train the velocity field at this interpolant to match the displacement from \( \bz(0)\) to \( {y} \) via the flow-matching loss (Equation~\ref{eq:standard_flow_matching_loss}):
% \begin{olivebox}
% \vspace{-0.4cm}
\begin{align}
\label{eq:final_floq_loss}
\mathcal{L}_{\methodname{}}(\theta) = \mathbb{E}_{\bz, t} \left[ \left\| v_\theta(t, \bz(t) \mid s, a) - \frac{({y}(s, a) - \bz)}{1 - 0} \right\|_2^2 \right].
\end{align}
\vspace{-0.25cm}
\subsection{{Preventing Flow Collapse: How to Make \methodname{} Work Well?}} 
\label{sec:practical}
\vspace{-0.15cm}

So far, we have introduced a conceptual recipe for parameterizing and training a Q-function critic via flow matching. However, a na\"ive instantiation of this idea performed no better than a standard monolithic Q-function in our initial experiments. This weak performance is the result of the inability of the network to meaningfully condition on the interpolant $\bz(t)$, causing the flow model to collapse to a monolithic Q-network. Interestingly, we find that this problem can be addressed by targeting two pathologies associated with applying flow-matching to TD: training with constantly evolving targets and running the flow on a scalar Q-values. 
We describe our approach for handling these pathologies, and to do so, we first answer: \emph{what constitutes a ``healthy''} \methodname{} \emph{velocity field?} Then we introduce two crucial modifications to the \methodname{} architecture that enable learning healthy \methodname{} networks. 

\begin{figure}[t]
\centering
\vspace{-0.2cm}
\includegraphics[width=0.99\textwidth]{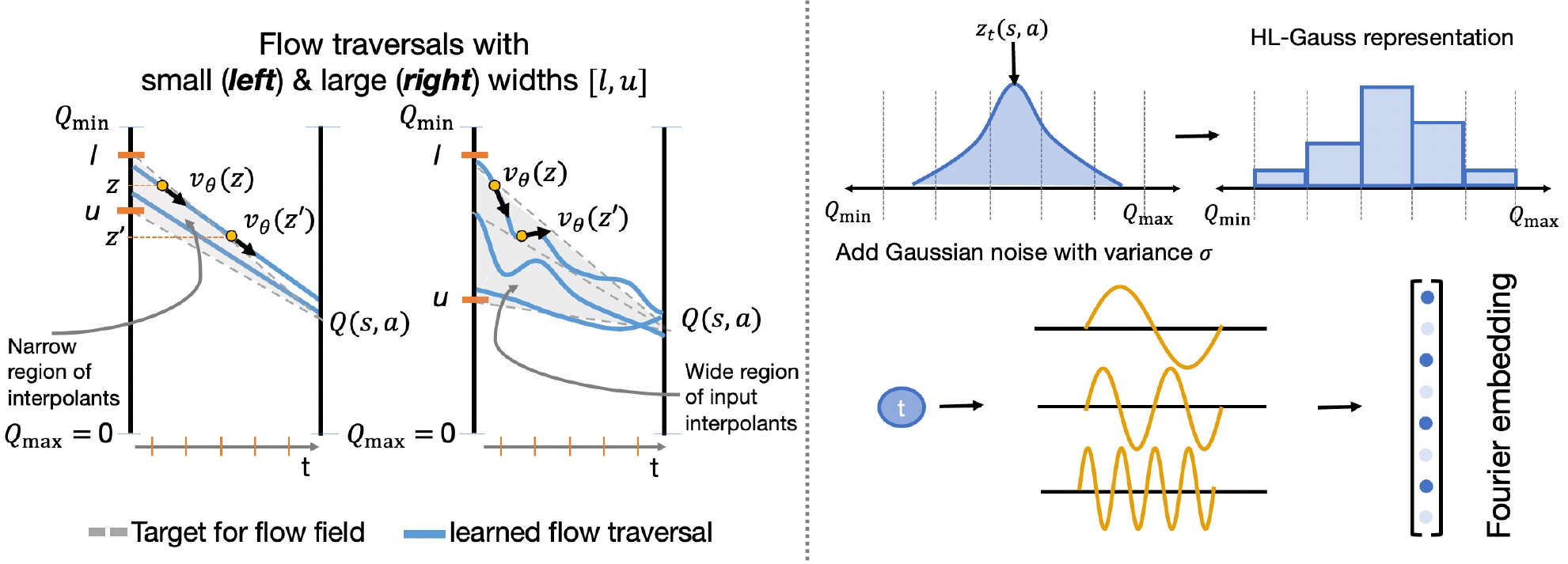}
\vspace{-0.4cm}
\caption{\footnotesize 
\emph{\textbf{Illustrating the role of our design choices.}} \textbf{Left:} When the width of the interval $[l, u]$ is small, and the overlap between this interval and the range of target Q-values we hope to see is minimal, we would expect to see more straight flow traversals, that might be independent of interpolant $\bz$. However, with wider intervals $[l, u]$, the flow traversal would depend on $\bz$, and hence span a curved path when running numerical integration during inference. \textbf{Right:} Illustrating how we transform an input interpolant $\bz$ into a categorical representation (top) and converting time $t$ into a Fourier-basis embedding (bottom). 
}
\label{fig:flow_illustration}
\vspace{-0.3cm}
\end{figure}

\textbf{\emph{When is} \methodname{} \emph{effective?}} Unlike traditional applications of flows, \methodname{} applies them to scalar Q-values. How does flow matching on a scalar work? Consider the trajectory traced by the flow during inference, as it evolves from initial noise ($t=0$) to the Q-value estimate produced by the network ($t=1$){ (Figure~\ref{fig:flow_illustration}; left)}. If this trajectory is a straight line, the velocity field $v_\theta(\bz(t),t)$ does not need to depend on $t$ and predicting a constant velocity proportional to the target Q-value is sufficient. In this case, flow matching provides no additional capacity beyond a monolithic Q-network. In contrast, if the trajectory is curved, the velocity field must utilize the interpolant $\bz(t)$ and time $t$ to predict customized velocities and be able to integrate to an accurate Q-value estimate at $t=1$. Thus, even though training uses a simple linear flow-matching loss, extra capacity emerges only when the learned flows produce (slightly) curved trajectories. In this regime, iterative computation amplifies model capacity, allowing \methodname{} to outperform monolithic Q-networks. 
We do remark that overly curved flows are also problematic as they amplify errors in the integration process itself. Therefore, our design choices do not explicitly increase how curved the flow is, but simply enable learning curved flow traversals when needed.

\textcolor{lightblue}{\textbf{{Design choice 1: Distribution of the initial noise sample.}}} As shown in prior works~\citep{flow_lipman2023, liu2023flow}, rescaling the source noise leaves the target distribution unchanged, but it alters the curvature of the transport trajectories. Interestingly, this effect seems to be particularly pronounced when applying flow-matching to scalar TD-learning (see Figure~\ref{fig:noise_ablation} in experiments). As such, we find that that setting the bounds $l$ and $u$ for the distribution of the initial noise, $\mathrm{Unif}~[l, u]$ greatly affects the performance of \methodname{}. 

We hypothesize that two aspects are important: (a) how close the target Q-values during training are to the chosen interval $[l, u]$, and (b) the width of the interval $u - l$. If the width $u - l$ is too small, then the interpolants $\bz(t)$ span only a very limited range of values. When we then run (imperfect) TD-loss training on these interpolants, the network parameterizing the velocity field receives little meaningful variation in $\bz(t)$ to associate changes in target values with. As a result, the model fails to exploit $\bz(t)$ effectively and degenerates into behaving like a standard monolithic Q-function. Likewise, if the interval $[l, u]$ is very disjoint from the range of target Q-values during training, then all interpolants $\bz(t)$ are forced to predict large velocities pointing in the general direction of the target Q-value. This reduces the need to learn calibrated velocity predictions conditioned on $\bz(t)$ and time $t$. We show this in Figure~\ref{fig:flow_illustration}, left.

We propose to choose $l$ and $u$ using a simple heuristic. Specifically, we set $u = Q_{\max}$ (=0 for most of our tasks). We then choose $l \geq Q_{\min}$ to maximize the interval width $u-l$, subject to stable learning curves and TD-error values comparable to a monolithic network. Here $Q_{\min}$ and $Q_{\max}$ denote the minimal and maximal possible Q-value achievable on the task. Thus, the interval $[l,u]$ is likely to overlap well with the target Q-values the velocity field must predict during training. We remark that this heuristic relies on the assumption that over the course of learning, Q-value predictions will evolve from near-zero values to the target value. Since standard network initializations already produce values near zero, this assumption is not limiting and helps cover the target Q-value range we hope to see.

\textcolor{lightblue}{\textbf{Design choice 2: Representing interpolant inputs to the velocity network.}} The second challenge stems from the fact that the magnitudes of the scalar interpolant $\bz(t)$ keep evolving throughout training. While standard TD-learning naturally must deal with non-stationary outputs, i.e., Q-values that increase from near-zero (due to random initialization) to larger magnitudes during training, this is usually not problematic with current best practices for TD-learning (that involve normalization of activations~\citep{nauman2024bigger}). In contrast, \methodname{} must deal with non-stationarity at the \emph{input} since the interpolant \( \bz(t) \), that changes over the course of learning is an input to the velocity flow. Indeed, we observe that as training progresses, the magnitude of \( \bz(t) \) grows, which can lead to high-magnitude gradients and be problematic.
%%AK.8.26: Can we add a figure here?

To address this, we adopt a categorical input representation of $\bz(t)$, inspired by the HL-Gauss encoding of \citet{farebrother2024stop}, which prevents the magnitude of the value $\bz(t)$ from substantially skewing activations in the velocity network and destabilizing training. To obtain this input representation, we first add Gaussian noise to \( \bz(t) \) with standard deviation \( \sigma \), then convert the resulting Gaussian PDF $\mathcal{N}(\bz(t), \sigma^2)$ into a categorical ``histogram'' distribution over \( N \) bins spanning the expected range of Q-values during training. Compared to \citet{farebrother2024stop}, we use a larger \( \sigma \), chosen such that approximately 80\% of the bins receive non-zero probability mass at initialization. This encourages broader coverage over bins and reduces sensitivity to non-stationarity. We also attempted to  rescale the interpolant to $[0, 1]$ and this did help to some extent, though we found that the HL-Gauss encoding of $\bz(t)$ performed best.

Finally, we also utilized a Fourier basis representation of the time variable $t$, provided as input to the velocity network $v_\theta(t, \bz(t))$. This is illustrated in Figure~\ref{fig:flow_illustration} (right). We show in our experiments than doing so helps substantially by encouraging the network to meaningfully utilize this time.

\begin{AIbox}{Summary: \methodname{} architecture and training}
\methodname{} parameterizes the Q-function via a learned velocity field. We train this velocity field with a linear flow-matching loss against the target Q-value (Eq.~\ref{eq:final_floq_loss}). To ensure that \methodname{} scales capacity, we carefully choose the distribution of the initial noise $\mathrm{Unif}~[l, u]$ to be wide and overlapping with range of target Q-values. We encode the interpolant using a categorical representation. We represent the input $t$ to $v_\theta$ using a Fourier encoding. See Algorithm~\ref{alg:cfm} for pseudocode and details.
\end{AIbox}

\vspace{-0.25cm}
\section{Experimental Evaluation}
\label{sec:exps}
\vspace{-0.25cm}

The goal of our experiments is to evaluate the efficacy of \methodname{} in improving offline RL and online fine-tuning. To this end, we compare \methodname{} to state-of-the-art methods, and answer the following questions: \textbf{(1)} Does \methodname{} improve performance when compared to using similar-sized networks on benchmark tasks? and 
\textbf{(2)} How does the use of iterative computation via \methodname{} compare with the use of ``parallel'' computation of a neural network ensemble and ``sequential'' iterative computation driven by ResNets of comparable size? We then run several experiments to understand the behavior of \methodname{} critics. Furthermore, we run a variety of ablation studies to understand the design choices that drive the efficient use of iterative computation, including the roles of \textbf{a)} tuning the width of initial noise sample, \textbf{b)} categorical representations of the interpolant input, and \textbf{c)} Fourier-basis time embeddings.

\vspace{-0.25cm}
\subsection{Main Offline RL Results}\label{sec:main_offline_rl}
\vspace{-0.15cm}
\begin{wrapfigure}{r}{0.5\textwidth}
\vspace{-0.7cm}
\centering
\includegraphics[width=0.99\linewidth]{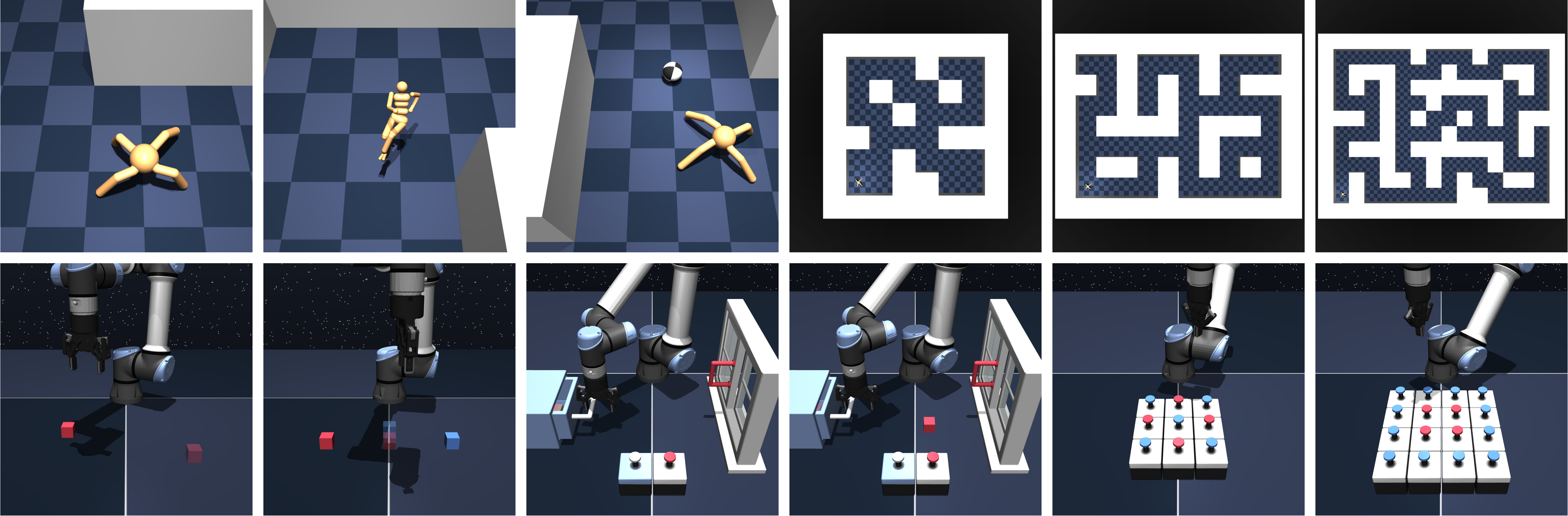}
\vspace{-0.6cm}
    \caption{\footnotesize \textbf{OGBench~\citep{ogbench_park2025} domains} that we study in this work. These tasks include high-dimensional state and action spaces, sparse rewards, stochasticity, as well as hierarchical structure.}   \label{fig:ogbench_tasks}
    \vspace{-0.4cm}
\end{wrapfigure}
\textbf{Offline RL tasks and datasets.} 
Following evaluation protocols from recent work in offline RL~\citep{park2025flow,wagenmaker2025steering,espinosa2025scaling}, we use the \mbox{\textbf{OGBench}} task suite~\citep{ogbench_park2025} as our main evaluation benchmark (see Figure \ref{fig:ogbench_tasks}).
OGBench provides a number of diverse, challenging tasks across robotic locomotion and manipulation,
where these tasks are generally more challenging than standard D4RL tasks~\citep{d4rl_fu2020}, which have been saturated as of 2024~\citep{rebrac_tarasov2023, d5rl_rafailov2024, park2024value}.
While OGBench was originally designed for benchmarking offline goal-conditioned RL,
we use its reward-based single-task variants (``\texttt{-singletask}'' from \citet{park2025flow}).
We employ $5$ locomotion %
and $5$ manipulation environments where each environment provides $5$ tasks, totaling to $\mathbf{50}$ state-based OGBench tasks. Some tasks are more challenging and longer-horizon (e.g., marked in the table).

\textbf{Comparisons and evaluation protocol.}
In addition to a \methodname{} critic, we build our experiments to utilize flow-matching policies. This implies that {flow-Q learning} (\textbf{FQL})~\citep{park2025flow}, which utilizes a flow-matching policy but with a monolithic Q-network, is our main comparison. The FQL work reports results with $1$M training steps; for a fair comparison, we additionally re-run FQL for $2$M steps and report both sets of results.  We run \methodname{} with a default set of hyperparameters across tasks, but on the more challenging \texttt{humanoidmaze-large} and \texttt{antmaze-giant} tasks we found a larger batch size of $512$ to be more effective, which we use for both FQL and \methodname{} on these environments. Beyond {FQL}, we compare against three recent state-of-the-art offline RL algorithms, all of which rely on monolithic Q-functions: (i) \textbf{ReBRAC}~\citep{rebrac_tarasov2023}, which corresponds to the strongest-performing method with a monolithic Q-network and a Gaussian policy, and thus provides a competitive non-flow baseline; (ii) \textbf{DSRL} \citep{wagenmaker2025steering} that adapts a diffusion-based behavior cloning policy by performing RL over its latent noise space, improving over FQL; 
% \nauman{we seem to have a bit inconsistent use of textbf in this paragraph}
; and (iii) \textbf{SORL} \citep{espinosa2025scaling} that leverages shortcut flow models to improve the training and inference scalability of flow policies. Note that none of them utilize a flow-matching Q-function, but do innovate across various properties of policy training. Comparing to the strongest methods that innovate on the design of the policy allows us to evaluate the importance of and benefits from utilizing flows for a critic.

\setlength{\tabcolsep}{3pt}
\begin{table*}[t!]
\vspace{-10pt}

\caption{
\footnotesize
\textbf{Offline RL results (all tasks).}
\methodname{} achieves competitive or superior performance compared to prior approaches. ``Hard'' environments refers to the set of environments where the FQL approach attains below 50\% performance, averaged over the $5$ tasks. \methodname{} is especially more performant on these hard environments over prior comparisons, where its performance (with best configuration) is around 1.8$\times$ of FQL. We don't report DSRL~\citep{wagenmaker2025steering} here as this prior work does not run on the exhaustive set of tasks (see Table~\ref{table:default_task_offline_results} for these). A comparison on just the default tasks reveals \methodname{} outperforms DSRL by >2$\times$.}

\label{table:all_task_offline_results}
\centering
\vspace{-0.1cm}
\resizebox{0.99\linewidth}{!}
{
\begin{threeparttable}
\begin{tabular}{l cc ccc | cc}

\toprule

\multicolumn{1}{c}{} & \multicolumn{2}{c}{\texttt{Gaussian Policy}} & \multicolumn{3}{c}{\texttt{Flow Policy}} & \multicolumn{2}{c}{\texttt{Flow Q-function (Ours)}} \\
\cmidrule(lr){2-3} \cmidrule(lr){4-6} \cmidrule(lr){7-8}
\texttt{Environment (5 tasks each)} & \texttt{BC} & \texttt{ReBRAC} & \texttt{SORL} &\texttt{FQL (1M)} & \texttt{FQL(2M)} &\texttt{\methodname{} (Def.)} &\texttt{\methodname{} (Best)} \\

\midrule

\texttt{antmaze-large}  &$11$ {\tiny $\pm 1$}& $81$ {\tiny $\pm 5$}  & $89$ {\tiny $\pm 2$} & $79$ {\tiny $\pm 3$}& $83$ {\tiny $\pm 5$}& \mathbf{$91$} {\tiny $\pm 5$} & \mathbf{$91$} {\tiny $\pm 5$}\\

\texttt{antmaze-giant} \textcolor{pink}{\textit{(Hard)}} &$0$ {\tiny $\pm 0$} & $26$ {\tiny $\pm 8$}  & $9$ {\tiny $\pm 6$} & $22$ {\tiny $\pm 19$} & $27$ {\tiny $\pm 23$}& $36$ {\tiny $\pm 21$} & \mathbf{$51$} {\tiny $\pm 12$}\\

\texttt{hmmaze-medium}  &$2$ {\tiny $\pm 1$}& $22$ {\tiny $\pm 8$}  & $64$ {\tiny $\pm 4$} & $57$ {\tiny $\pm 5$}& $69$ {\tiny $\pm 20$}& \mathbf{$82$} {\tiny $\pm 10$} & \mathbf{$82$} {\tiny $\pm 10$}\\

\texttt{hmmaze-large} \textcolor{pink}{\textit{(Hard)}} &$1$ {\tiny $\pm 0$}& $2$ {\tiny $\pm 1$}  & $5$ {\tiny $\pm 2$} & $9$ {\tiny $\pm 6$}& $16$ {\tiny $\pm 9$}& \mathbf{$28$} {\tiny $\pm 9$} & \mathbf{$28$} {\tiny $\pm 9$}\\

\texttt{antsoccer-arena}  &$1$ {\tiny $\pm 0$} & $0$ {\tiny $\pm 0$}  & \mathbf{$69$} {\tiny $\pm 2$} & $60$ {\tiny $\pm 2$}& $61$ {\tiny $\pm 10$}& $65$ {\tiny $\pm 12$} & $65$ {\tiny $\pm 12$}\\

\texttt{cube-single}  &$5$ {\tiny $\pm 1$}& $91$ {\tiny $\pm 2$}  & $97$ {\tiny $\pm 1$}& $96$ {\tiny $\pm 1$}& $94$ {\tiny $\pm 5$}& \mathbf{$98$} {\tiny $\pm 3$} & \mathbf{$98$} {\tiny $\pm 3$}\\

\texttt{cube-double} \textcolor{pink}{\textit{(Hard)}}  &$2$ {\tiny $\pm 1$}& $12$ {\tiny $\pm 1$} & $25$ {\tiny $\pm 3$} & $29$ {\tiny $\pm 2$}& $25$ {\tiny $\pm 6$}& \mathbf{$47$} {\tiny $\pm 15$} & \mathbf{$47$} {\tiny $\pm 15$}\\

\texttt{scene} &$5$ {\tiny $\pm 1$}& $41$ {\tiny $\pm 3$}  & $57$ {\tiny $\pm 2$} & $56$ {\tiny $\pm 2$}& $57$ {\tiny $\pm 4$}& \mathbf{$58$} {\tiny $\pm 6$} & \mathbf{$58$} {\tiny $\pm 6$}\\

\texttt{puzzle-3x3} \textcolor{pink}{\textit{(Hard)}} &$2$ {\tiny $\pm 0$}& $21$ {\tiny $\pm 1$}& $-$ {\tiny $\pm -$}& $30$ {\tiny $\pm 1$}& $29$ {\tiny $\pm 5$}& \mathbf{$37$} {\tiny $\pm 7$} & \mathbf{$37$} {\tiny $\pm 7$}\\

% \texttt{puzzle-4x4} \textcolor{pink}{\textit{(Hard)}} &$0$ {\tiny $\pm 0$} & $14$ {\tiny $\pm 1$} & $-$ {\tiny $\pm -$}& $17$ {\tiny $\pm 2$}& $9$ {\tiny $\pm 3$}& $21$ {\tiny $\pm 5$ } & \mathbf{$28$} {\tiny $\pm 6$ }\\
\texttt{puzzle-4x4} \textcolor{pink}{\textit{(Hard)}} &$0$ {\tiny $\pm 0$} & $14$ {\tiny $\pm 1$} & $-$ {\tiny $\pm -$}& $17$ {\tiny $\pm 2$}& $9$ {\tiny $\pm 3$}& $21$ {\tiny $\pm 5$ } & \mathbf{$28$} {\tiny $\pm 6$ }\\

\midrule

\texttt{Average Score (All Environments)} & $3$ & $31$ & $-$ & $46$ & $47$ & $56$ & $\mathbf{59}$\\
\texttt{Average Score (Hard Environments)} & $1$ & $15$ & $-$ & $21$ & $21$ & $34$ & $\mathbf{38}$ \\

\bottomrule
\end{tabular}

\end{threeparttable}
}
\end{table*}

\textbf{\methodname{} configuration.} We report results for \methodname{} with two configurations. The ``default'' configuration utilizes the same hyperparameters across tasks and environments, whereas the ``best'' configuration uses environment-specific hyperparameters that still are fixed across all tasks in that environment, to get a sense of the upper bound with \methodname{}. However, as we will show, even the default configuration of \methodname{} substantially outperforms all prior methods. The default configuration utilizes $K = 8$ flow steps and sets the width of $u - l$ to be $\kappa \times (Q_{\max} - Q_{\min})$, where $\kappa = 0.1$. The best configuration tunes the number of flow steps $K \in \{4,8\}$ and $\kappa \in \{0.1, 0.25\}$ per environment (\emph{not} per task). For a fair comparison with FQL, we use a $4$-layer flow critic in all environments, except in the cube (single and double) environments where we employ a smaller $2$-layer flow critic because we saw training instabilities with 4-layer critics.

\begin{figure}[t]
    \centering
    \vspace{-0.2cm}
    \includegraphics[width=0.99\textwidth]{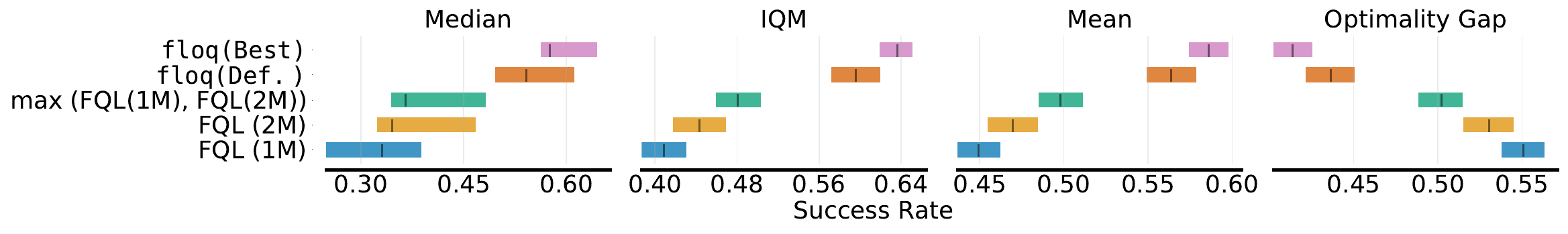}
    \vspace{-0.3cm}
    \caption{\footnotesize Comparison of \methodname{} against the baseline FQL across median, interquartile mean (IQM), mean and optimality gap, following \citet{rliable_agarwal2021}. Results show that \methodname{} consistently outperform FQL across all evaluation criteria with no confidence interval overlap in all cases, meaning that the gains from \methodname{} are significant.}
    \label{fig:rliable_1}
    \vspace{-0.4cm}
\end{figure}

\textcolor{lightblue}{\textbf{Empirical results.}} Observe in Table~\ref{table:all_task_offline_results} that \methodname{} outperforms prior methods, including FQL, on average across all 50 tasks, evaluated over 3 seeds for each task. Also note that \methodname{} improves over FQL most on the harder environments, where FQL attains performance below $50\%$ success rate (antmaze-giant, hmmaze-large, cube-double, puzzle-3x3, and puzzle-4x4). For a statistically rigorous evaluation, we adopt techniques from \citet{rliable_agarwal2021} and plot various statistics of the comparisons between \methodname{} and FQL: \textbf{1)} median and IQM scores across the 50 tasks in Figure~\ref{fig:rliable_1}, where we do not observe \emph{any} overlap between the confidence intervals; \textbf{2)} performance profile for \methodname{} in Figure~\ref{fig:rliable_2}, which we see is strictly superior for \methodname{}; and \textbf{3)} the $P(X>Y)$ statistic in Figure~\ref{fig:rliable_2}, which also confirms the efficacy of \methodname{}. 

Since DSRL~\citep{wagenmaker2025steering} only evaluates on the 10 default tasks (one default task per environment), we also provide an additional results table on only the default tasks in each environment in the appendix (Table~\ref{table:default_task_offline_results}). On the default tasks, we find that \methodname{} outperforms DSRL (20\% for DSRL vs. 45\% for \methodname{}), improving by over $2 \times$ in success rate. While \methodname{} uses an expected Q-value backup it still learns a stochastic Q-function. Thus, we also compare \methodname{} to a representative distributional RL approach (IQN~\citep{dabney2017distributional}) in Table \ref{table:default_task_offline_results}, and we observe that \methodname{} outperforms IQN. These results establish the efficacy of \methodname{}, and generally, performance improvements in offline RL from scaling Q-functions via flow-matching models.

\vspace{-0.25cm}
\subsection{Main Online Fine-Tuning Results}
\vspace{-0.2cm}
\begin{figure}[t]
\centering
\vspace{0.2cm}
\includegraphics[width=0.99\textwidth]{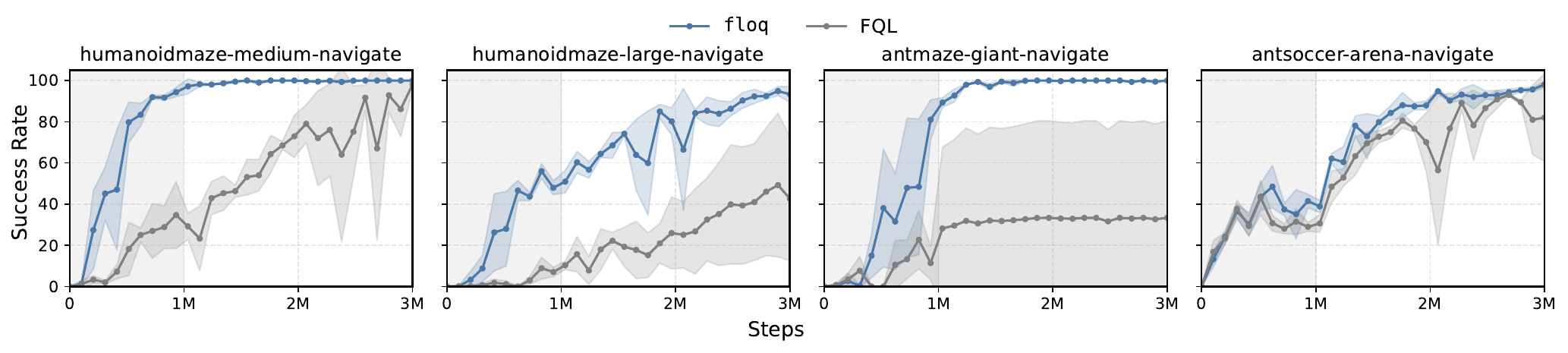}
\vspace{-0.4cm}
\caption{\footnotesize \emph{\textbf{Learning curves for online fine-tuning}} of \methodname{} and FQL across four hard tasks. \methodname{} not only provides a stronger initialization from offline RL training but also maintains its advantage throughout online fine-tuning on the hardest tasks, leading to faster adaptation and higher final success rates. The shaded gray area denotes offline RL training.
}
\vspace{-0.3cm}
\label{fig:online_finetuning}
\end{figure}
Next, we evaluate the efficacy of \methodname{} in the online RL fine-tuning setting. Here, we first train agents completely offline for $1M$ steps and subsequently fine-tune them online for an additional $2M$ steps (Figure~\ref{fig:online_finetuning}). 
% We set the updates-to-data (UTD) ratio to be equal to 1.0 for both the base FQL approach and \methodname{}.
Across four challenging tasks ({humanoidmaze-medium}, {humanoidmaze-large}, {antmaze-giant}, and {antsoccer-arena}), we find that \methodname{} provides a substantially stronger initialization compared to FQL and also adapts more rapidly during online interaction and converges to higher final performance. In contrast, FQL suffers from poorer initializations and exhibits slower rates of policy improvement. In the complete set of results in Figure~\ref{fig:online_finetuning_all}, we again see that \methodname{} matches or outperforms FQL on each task.

\begin{figure}[t]
\centering
\vspace{-0.2cm}
\includegraphics[width=0.99\textwidth]{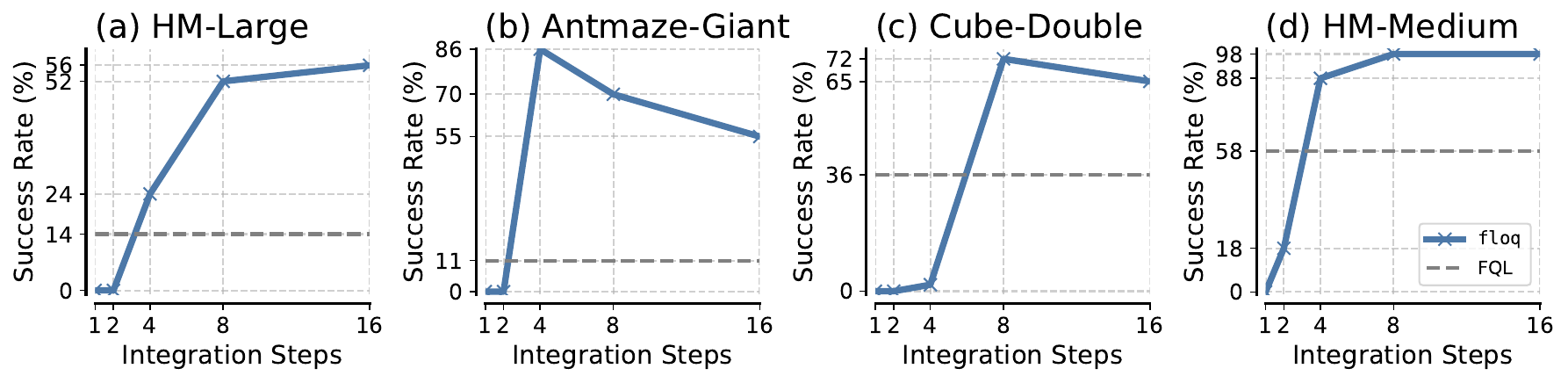}
\vspace{-0.4cm}
\caption{\footnotesize 
\emph{\textbf{Effect of integration steps on \methodname{} performance.}} Performance of \methodname{} with varying flow steps, compared against a monolithic Q-function (FQL). We run these experiments with the default configuration for \methodname{}. More flow steps generally improve performance, but too many steps can lead to diminishing or negative returns (e.g., \texttt{antmaze-giant}). That said, in all configurations, \methodname{} outperforms FQL, and utilizing a moderately large number of flow steps is important.
}
\label{fig:flow_steps_ablations}
\vspace{-0.3cm}
\end{figure}

\vspace{-0.25cm}
\subsection{Understanding the Scaling Properties and Behavior of \methodname{}}
\label{sec:understanding_behavior}
\vspace{-0.15cm}

To better understand the benefits of iterative compute in \methodname{}, we analyze its scaling behavior and compare it to different scaling approaches. Specifically, we study: \textbf{(i)} how the number of flow-integration steps controls the expressivity of \methodname{}; \textbf{(ii)} how \methodname{}’s iterative computation compares to monolithic critic scaling; and \textbf{(iii)} the importance of applying supervision to the velocity field at every flow step.

\begin{wrapfigure}{r}{0.6\textwidth}
    \vspace{-0.3cm}
    \centering
    \includegraphics[width=0.97\linewidth]{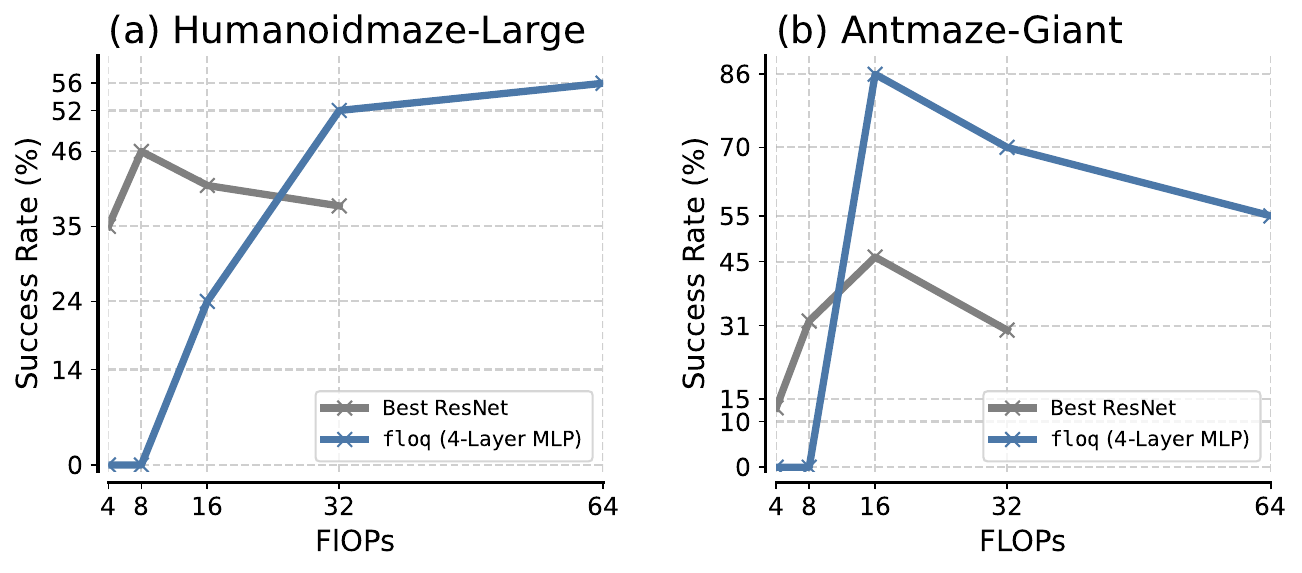}
    \vspace{-0.3cm}
    \caption{\footnotesize \emph{\textbf{Comparison of \methodname{} with ResNet critics on the hardest tasks}}: \texttt{hmmaze-large}, \texttt{antmaze-giant}. A $4$-layer flow critic outperforms the best ResNets under a total budget on forward pass capacity, even after tuning over multiple residual configurations. For any given value on the x-axis, we plot the performance of the best performing ResNet configuration at that inference cost. For \methodname{}, we run more integration steps at inference. Thus, our approach of training \methodname{} does not simply add more depth.}
    \label{fig:resnet_comparison}
    \vspace{-0.5cm}
\end{wrapfigure}

\textcolor{lightblue}{\emph{\textbf{1) How does the performance of}} \textbf{\methodname{}} \textbf{\emph{depend on the number of integration steps for the flow?}}}
We now study the effect of varying the number of integration steps for the flow in \methodname{}. In Figure \ref{fig:flow_steps_ablations}, we report the success rate of \methodname{} with $K \in \{1,2,4,8,16\}$ flow steps, alongside a monolithic Q-function (FQL) using the same architecture. Across environments, increasing the number of flow steps generally improves the performance of \methodname{}, with notable gains on harder tasks such as \texttt{hmmaze-large} and \texttt{antmaze-giant}. Importantly, even with as few as $4$ flow steps, \methodname{} already outperforms FQL in most settings, and the gap widens further with additional steps. However, we also observe diminishing returns and, in some cases, slight degradation beyond a moderate number of steps (e.g., on \texttt{antmaze-giant}, where $8$ and $16$ steps perform worse than $4$). We suspect that this degradation stems from overfitting when the number of integration steps for computing the target is excessive, and often manifests as unstable dynamics of TD-errors over the course of training. A similar degradation is observed for ResNets in Figure~\ref{fig:resnet_comparison}.

\begin{figure}[t]
\centering
\vspace{0.05cm}
\includegraphics[width=0.99\textwidth]{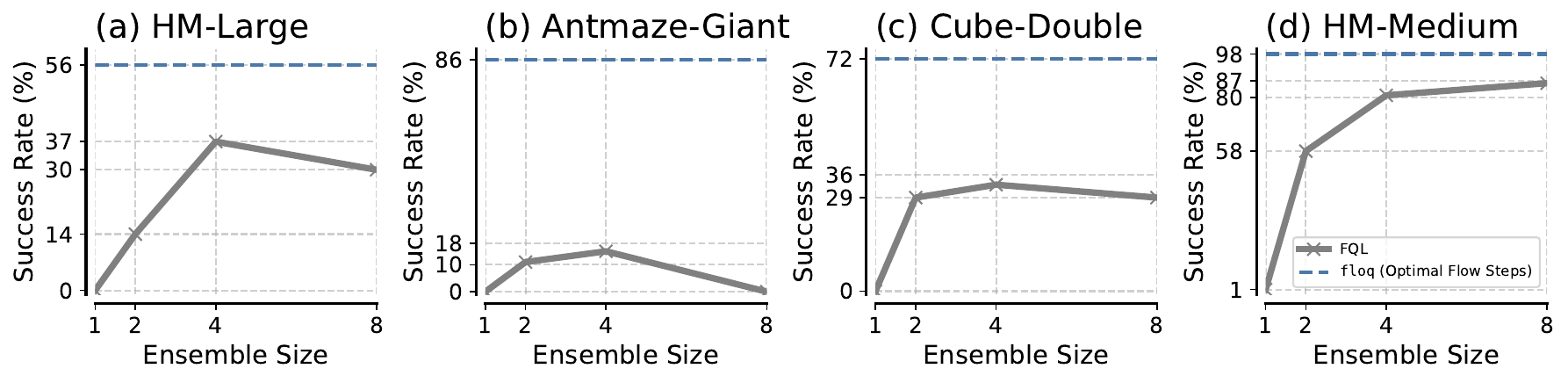}
\vspace{-0.4cm}
\caption{\footnotesize \textbf{Comparison of \methodname{} with monolithic ensembles.} We evaluate ensembles of size $1$, $2$, $4$, and $8$ using a monolithic critic of the same size as FQL. While larger ensembles improve FQL, even an $8$-critic ensemble falls short of \methodname{} with the same number of flow steps, showing that flow critics provide gains beyond parallel compute even in an compute-matched setting. 
}
\vspace{-0.3cm}
\label{fig:ensemble_ablations}
\end{figure}

\textcolor{lightblue}{\emph{\textbf{2) How does \methodname{} compare against increasing ``parallel'' compute of monolithic critics using ensembles?}}}
A natural way to increase capacity of monolithic Q-functions is through ensembling, i.e., using multiple Q-functions and averaging their predictions to compute the Q-value used for backups and policy updates. Doing so would increase parallel compute rather than sequential ``depth'' of the model, and can be parallelizable/preferable if it works well. We trained ensembles of $1$, $2$, $4$, and $8$ monolithic critics of the same size as the base FQL critic, and compared their performance against \methodname{}. In Figure~\ref{fig:ensemble_ablations}, we find that larger ensembles provide modest improvements over FQL performance, indicating improvements from scaling parallel compute in RL. However, even an ensemble of $8$ critics fails to match the performance of \methodname{}. This is despite the fact that \methodname{} runs $8$ steps of integration, meaning that it spends the same total compute (i.e., $8\times$ forward passes of a 4-layer MLP) for computing the Q-value as an ensemble of size 8. Thus, benefits of flow critics cannot be obtained simply by averaging multiple monolithic models, and instead arise from the sequential integration process inherent to the flow parameterization.

\textcolor{lightblue}{\emph{\textbf{3) How does \methodname{} compare against increasing ``sequential'' compute of monolithic critics?}}} Another hypothesis is that \methodname{} could be implementing a similar strategy as architectures like ResNet~\citep{resnet} that already perform some form of iterative computation. ResNet and \methodname{} are different due to the presence of dense supervision after each computation block in \methodname{}, which is absent in ResNets after each residual block since all supervision comes from the loss on the final prediction. To test whether this matters, we compare \methodname{} to a ResNet. Specifically, we use the same $4$-layer flow critic from before and benchmark it against the best-performing ResNets with FQL. We ran a ``cross-product'' over possible ResNet configurations with various block sizes ($2,4,8,16$ layers) and blocks ($8,4,2$). These configurations represent all ways to build a ResNet where  32 layers are involved in a forward pass.

We compare \methodname{} to the best performing ResNet under a given total inference compute budget (i.e., given an upper bound on number of feed-forward layers used during a forward pass) on {humanoidmaze-large} and {antmaze-giant}, since these hard environments should benefit from the more nuanced ResNet architectures the most. Observe in Figure~\ref{fig:resnet_comparison}, while ResNet critics do improve over base FQL, they still remain worse than \methodname{}, even under matched inference compute. Note that \methodname{} does not itself utilize a ResNet architecture (though its velocity network could use residual layers). Also note that while ResNet architectures instantiate a new set of parameters for every new layer added into the network, \methodname{} still only utilizes parameters from a \emph{single} 4-layer MLP for all values of inference capacity. This indicates that the gains of flow critics cannot be attributed to the phenomenon of adding more residual layers or more parameters, but rather the dense supervision provided by supervising the velocity field at each step of iterative computation. It also indicates that flow critics can perform better by scaling test-time compute without any extra parameters.

\textcolor{lightblue}{\emph{\textbf{4) Does \methodname{} benefit from densely supervising the velocity at all time steps?}}}
To answer this question, we trained a variant of \methodname{} where we supervised the velocity field was supervised only at $t=0$ and used only a single flow step for integration. Note that while one might think that training \methodname{} at $t=0$ is equivalent to baseline FQL, we clarify this is not the case. This is because while we do utilize only a single integration step, the input to the velocity network is still a randomly sampled scalar noise. The velocity field is trained to predict the difference between the target Q-value and this input noise value, \emph{for all different values of this noise}. This presents several ``auxiliary tasks'' for fitting the Q-function as opposed to just one in baseline FQL (i.e., the noise is set to $0$ only). We hypothesize training the network to fit these auxiliary tasks does result in representational benefits, consistent with conventional wisdom in TD-learning that relates auxiliary losses with representational benefits~\citep{lyle2021effect}. 

\begin{wrapfigure}{r}{0.6\textwidth}
  \centering
  \vspace{-0.4cm}
  \includegraphics[width=0.995\linewidth]{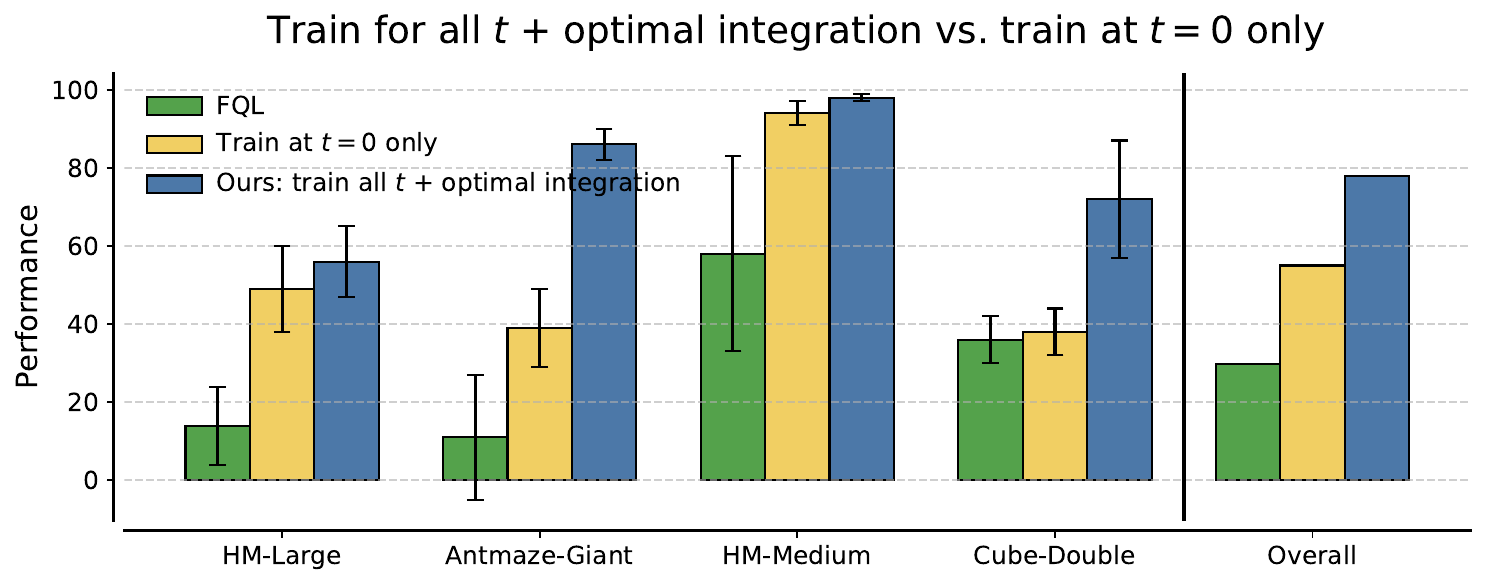}
  \vspace{-0.45cm}
  \caption{\footnotesize \emph{\textbf{Performance comparison between FQL, training \methodname{} only at $t=0$, 
  and the full \methodname{}} approach} with supervision across all $t$ (and optimal integration steps $k$ chosen from $\{4,8,16\}$). While $t=0$ training 
  improves over FQL, full \methodname{} consistently achieves the best performance, showing the benefits from training at all integration steps.}
  \label{fig:train_at_zero_only}
  \vspace{-0.4cm}
\end{wrapfigure}

As shown in Figure \ref{fig:train_at_zero_only}, this restricted variant substantially outperforms FQL but consistently underperforms the full \methodname{}, which supervises velocity at all $t \in [0,1]$ and leverages multiple integration steps. For example, on the {humanoidmaze-large} environment, performance increases from 14\% (FQL) to 49\% with only $t=0$ training of \methodname{}, but full \methodname{} achieves 56\%. On {antmaze-giant}, the gap is more pronounced, 
with scores of 11\%, 39\%, and 86\%, respectively. We observed similar patterns on {hm-medium} (58\%, 94\%, 98\%) and {cube-double} (36\%, 38\%, 72\%). We hypothesize that the gain from multiple flow steps on {hm-medium} is smaller because it is a simpler environment. These results suggest that while restricting supervision to $t=0$ already brings notable representation learning benefits, supervising at all $t$ and multi-step integration is important for unlocking the full potential of \methodname{}.

\begin{AIbox}{Takeaways: Properties and Behavior of \methodname{}}
\begin{itemize}[itemsep=2pt]
 \setlength{\leftskip}{-20pt}
    \item More integration steps are better, but performance saturates and can degrade at very high values.
    \item \methodname{} outperforms approaches that increase compute for monolithic Q-functions, whether through parallel ensembling or sequential depth expansion via ResNets.
    \item \methodname{} enhances representation learning, outperforming monolithic Q-functions even with just one integration step, though multiple integration steps are required for best performance.
\end{itemize}
\end{AIbox}

\vspace{-0.25cm}
\subsection{Ablation Studies for \methodname{}}
\vspace{-0.15cm}
Finally, in this section, we present experiments ablating various design choices and hyperparameters in \methodname{}. Our goal is to evaluate the sensitivity of \methodname{} to these choices and prescribe thumb rules for tuning them. Concretely, the design choices we ablate in this section include: \textbf{a)} the approach for embedding the interpolant $\bz(t)$, \textbf{b)} the approach for embedding ``time'' of the flow step $t$, and \textbf{c)} the range $[l, u]$ that provides the support for the initial noise sample $\bz(0)$.
 
\begin{figure}[t]
    \vspace{-0.2cm}
    \centering
    \includegraphics[width=0.99\linewidth]{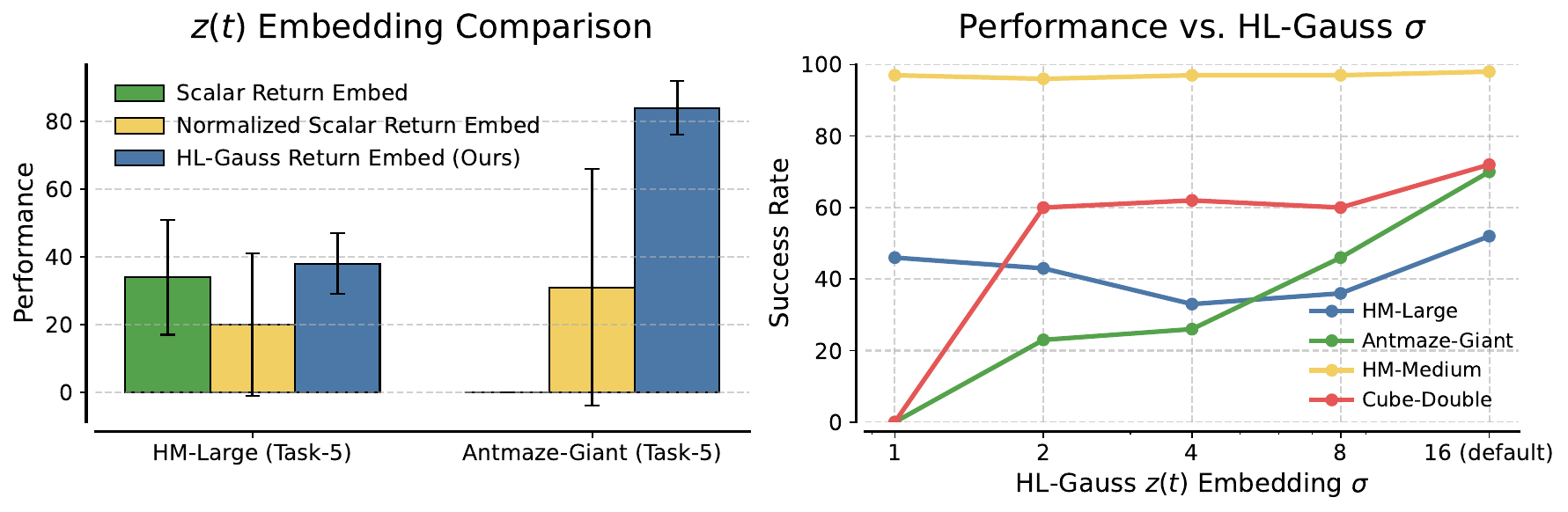}
    \vspace{-0.4cm}
    \caption{\footnotesize \emph{\textbf{Comparison of different approaches for representing the input interpolant in \methodname{}}}. \textbf{Left}: performance on two representative tasks where that HL-Gauss embeddings outperform scalar and normalized scalar embeddings by reducing sensitivity to non-stationary inputs. \textbf{Right:} Ablation over HL-Gauss embedding scale $\sigma$ for the scalar flow interpolant input, showing that larger values provide broader bin coverage and stronger performance. Default $\sigma = 16.0$.}
    \label{fig:hl_gauss_sigma}
    \vspace{-0.4cm}
\end{figure}

\textcolor{lightblue}{\emph{\textbf{1) How does the approach of embedding the interpolant $\bz(t)$ affect \methodname{} performance?}}} 
We observe that the approach of embedding $\bz(t)$ (Design Choice 2 in Section~\ref{sec:practical}) plays a significant role in the performance of \methodname{}. As shown in Figure~\ref{fig:hl_gauss_sigma}, HL-Gauss embeddings of $\bz(t)$ provide a significant advantage over scalar or normalized scalar embeddings. In particular, across several tasks we found HL-Gauss embeddings (with a sufficiently large value of $\sigma$) to be essential for achieving strong performance, and Figure~\ref{fig:hl_gauss_sigma} (left) highlights two representative tasks in this category. 
HL-Gauss embeddings with broader bin coverage helps reduce the sensitivity of the network to non-stationary inputs, thereby stabilizing training and improving performance. While normalizing $\bz(t)$ helps on some tasks over using the raw value, we found that HL-Gauss embeddings generally gave the best performance. In our implementation of the velocity network, we use HL-Gauss embeddings with a default scale of $\sigma = 16.0$. Figure~\ref{fig:hl_gauss_sigma} (right) shows an ablation over smaller values of $\sigma \in \{1.0, 2.0, 4.0, 8.0\}$. Observe that larger values of $\sigma$ consistently yield stronger performance. Intuitively, increasing $\sigma$ leads to broader bin coverage for the HL-Gauss distribution (see Figure~\ref{fig:flow_illustration}, right), which helps mitigate the non-stationarity of the range of $\bz(t)$ over the course of training with TD-learning.  These results highlight that selecting sufficiently large embedding scales is important for stabilizing learning and achieving strong downstream performance.

\begin{wrapfigure}{r}{0.6\textwidth}
    \vspace{-0.3cm}
    \centering
    \includegraphics[width=0.99\linewidth]{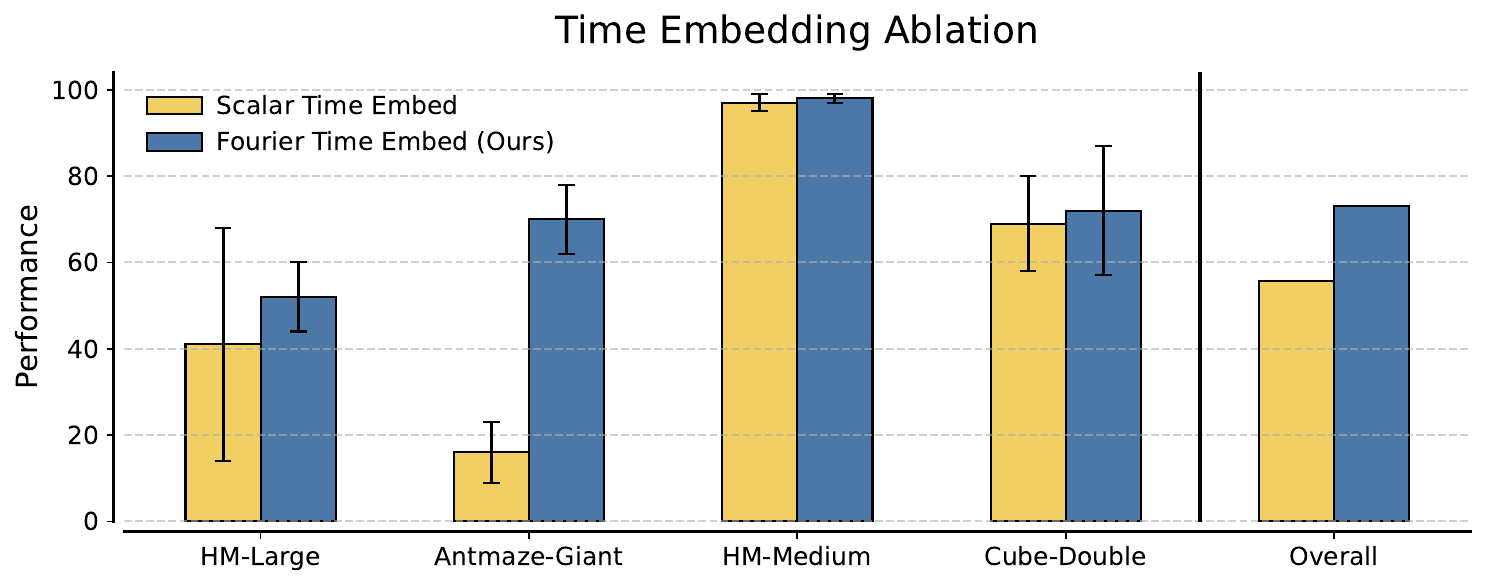}
    \vspace{-0.3cm}
    \caption{\footnotesize \emph{\textbf{Time embedding.}} Replacing the Fourier-basis embedding of time with a scalar embedding results in significantly worse performance, highlighting the importance of Fourier features for conditioning on time.}
    \label{fig:time_embedding_ablation}
    \vspace{-0.4cm}
\end{wrapfigure}
\textcolor{lightblue}{\emph{\textbf{2) Ablations for the time embedding.}}}
In the default configuration of \methodname{}, we used a 64-dimensional Fourier embedding for the time $t$, provided as input to the velocity field (also see Dasari et. al \cite{dasari2024ingredients} for a recent work training a diffusion policy also using Fourier embedding of $t$). As shown in Figure \ref{fig:time_embedding_ablation}, replacing this Fourier embedding with a simple scalar embedding of $t$ leads to a significant drop in performance on several tasks. This highlights the importance of the Fourier embedding, which allow the velocity function to be meaningfully conditioned on $t$, enabling it to produce distinct behaviors at different integration times. Without such rich embeddings, the critic struggles to leverage temporal information effectively, and again collapse to the monolithic architecture. We therefore recommend that practitioners carefully utilize high-dimensional embeddings of time when using \methodname{}.

\begin{figure}[t]
    % \vspace{-0.5cm}
    \centering
    \includegraphics[width=0.85\linewidth]{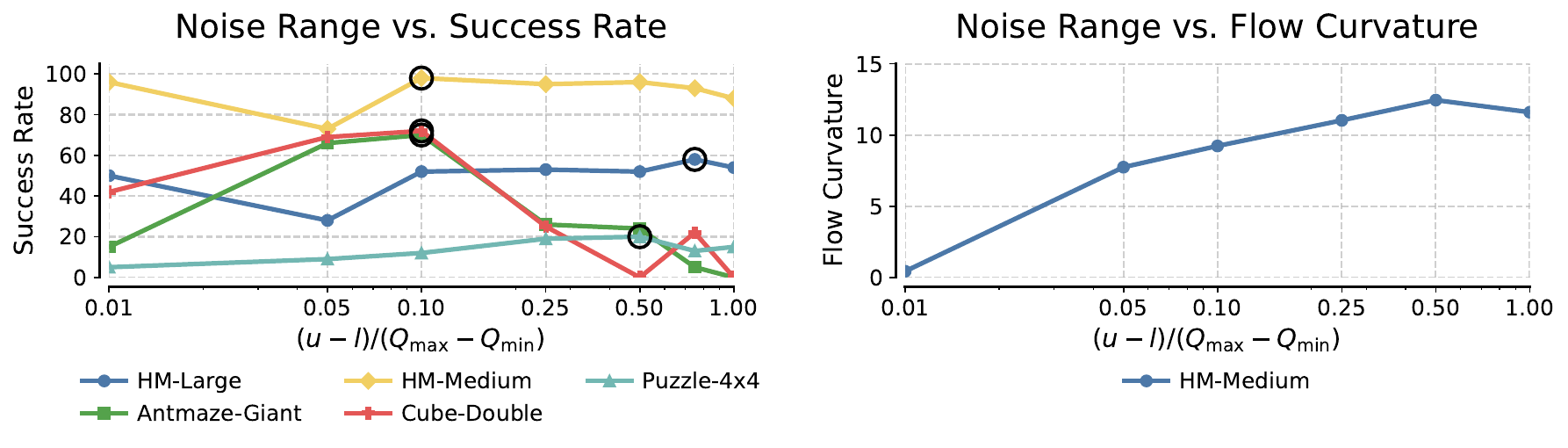}
    \vspace{-0.4cm}
    \caption{\footnotesize 
    \emph{\textbf{Effect of variance of the initial noise sampling distribution on \methodname{}.}} 
    \textbf{Left:} Success rates across environments as a function of the initial noise scaling factor (black circles denote the best setting per environment). 
    \textbf{Right:} Flow curvature in HM-Medium increases with noise variance, highlighting the tradeoff between too little curvature (flow collapses to monolithic critic) and too much curvature (difficult numerical integration).}
    \label{fig:noise_ablation}
    % \vspace{-0.1cm}
\end{figure}

\textcolor{lightblue}{\textbf{3) Ablations for the width of the $[l, u]$ interval.}}
Finally, we study the effect of varying the variance of the initial noise sample used in critic flow matching by expanding the width $u-l$ of the interval 
that the initial noise is sampled from. We present the results in Figure~\ref{fig:noise_ablation}. On the left, we observe that the  performance across several tasks typically peaks at intermediate variance values (note that the black circles marking the setting that yields the best success rate for each environment). This means that choosing an interval $[l, u]$ with a non-trivial width is important. As discussed in Section~\ref{sec:practical}, 
Figure~\ref{fig:noise_ablation} (right) shows that the curvature of the learned flow increases as the width of the interval grows. We measure curvature by computing the magnitude of the derivative of the velocity field as a function of time using finite differences. That is, concretely, we measured the expected value of $|\nicefrac{d v_{\theta}(t, \bz(t))}{dt}|$ 
across state-action pairs in the offline dataset and averaged this metric through training. 

Putting results in Figure~\ref{fig:noise_ablation} together, we note that some degree of curvature is necessary for best performance, which is expected because otherwise, the flow collapses to behave like a monolithic critic. That said, excessive curvature makes the flow numerically harder to integrate, ultimately degrading performance. Based on these observations, we recommend practitioners use $\kappa := (u-l)/(Q_{\max}- Q_{\min})$ in the range of $\{0.1,0.25\}$ as reliable starting points when tuning \methodname{} critic.

\begin{figure}[t]
    \vspace{-0.2cm}
    \centering
    \includegraphics[width=0.9\linewidth]{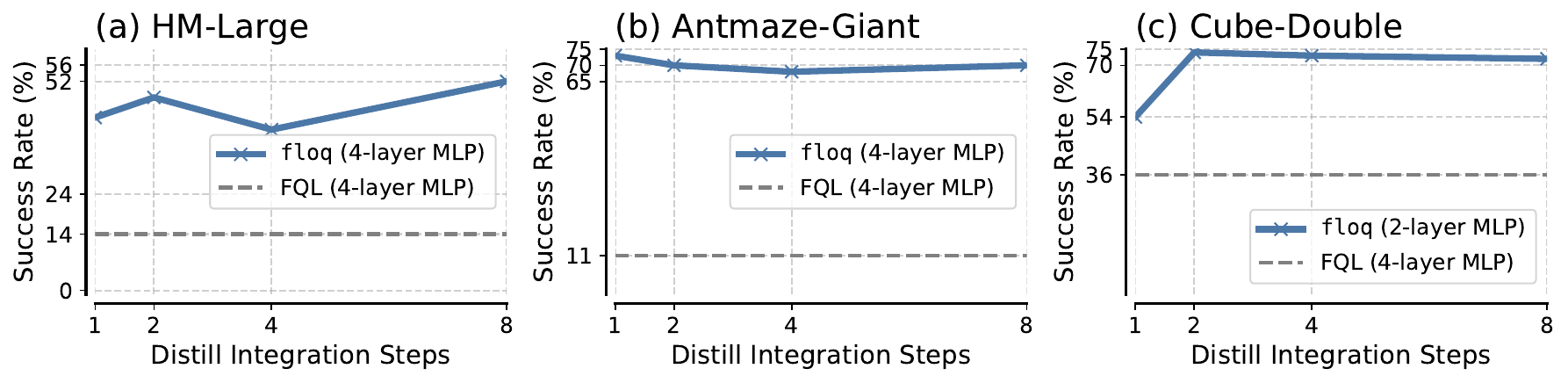}
    \vspace{-0.3cm}
    \caption{\footnotesize 
    \emph{\textbf{Effect of the number of integration steps used for policy extraction on \methodname{} performance.}} 
    Even though computing the target values for TD-learning utilizes a fixed number of $8$ integration steps, in this ablation we utilize a smaller number of steps for extracting the policy.
    Performance is more robust to the number of integration steps used for policy extraction, suggesting that as long as target integration is sufficiently accurate, few steps suffice for policy distillation.}
    \label{fig:distill_steps}
    \vspace{-0.4cm}
\end{figure}

\textcolor{lightblue}{\emph{\textbf{4) How does the number of critic flow steps used for the policy update affect the performance of \methodname{}?}}}
We next investigate the effect of varying the number of integration steps used for calculating the Q-value for the policy update. Since we build our algorithm on top of FQL, we implement the policy update by first distilling the values produced by the flow critic into a one-step, monolithic Q-function. Then the policy extraction procedure (akin to SAC+BC) maximize the values of this distilled critic subject to a behavioral cloning loss. Note that this approach essentially decouples the number of integration steps used to compute the TD-target and the number of integration steps for policy extraction.  As shown in Figure~\ref{fig:distill_steps}, as long as the number of integration steps for computing the target value are fixed (to $8$ in this case), the performance of \methodname{} is relatively robust to the number of integration steps used for the policy update. Contrast this with the sensitivity to the number of integration steps used for computing the TD-target observed in Figure~\ref{fig:flow_steps_ablations}. The results indicate that once the target integration steps are sufficiently large (here, 8), the policy can be effectively distilled even with a small number of integration steps.

\begin{AIbox}{Takeaways: Ablation studies for \methodname{}}
1) Utilizing an HL-Gauss embedding for $\bz(t)$ is crucial. Generally, the larger the coverage over bins, the better the performance of \methodname{} (Figure \ref{fig:hl_gauss_sigma}). 2) Utilizing a Fourier-basis embedding of time is critical for meaningfully conditioning on it (Figure \ref{fig:time_embedding_ablation}). 3) A moderate width of the initial noise distribution improves flow curvature, and performs best (Figure \ref{fig:noise_ablation}).
\end{AIbox}

\vspace{-0.2cm}
\section{Discussion and Perspectives on Future Work}
\vspace{-0.2cm}
In this paper, we presented \methodname{}, an approach for training critics in RL using flow-matching. \methodname{} formulates value learning as transforming noise into the value function via integration of a learned velocity field. This formulation enables scaling Q-function capacity by utilizing more compute during the process of integration to compute the Q-function. As a result of utilizing a flow-matching objective for training, \methodname{} utilizes dense supervision at every step of the integration process. We describe some important design choices to train flow-matching critics to make meaningful use of integration steps. Through our experiments, we show that \methodname{} attains state-of-the-art results on a suite of commonly-used offline RL tasks, and outperforms other ways of expanding capacity of a Q-function (e.g., via a ResNet or monolithic Q-function ensemble). We also show the necessity of learning curved flow traversals to make effective use of capacity and utilizing the design choices we prescribe in this work. 

\textbf{Future work.} We believe \methodname{} presents an exciting approach to scale Q-function capacity. Thus, there are a number of both theoretical and empirical open questions. From an empirical standpoint, it is important to understand how to appropriately set the number of integration steps as excessive steps may degrade Q-function quality. This degradation, however, is not localized to just flows but also to ResNets (Figure~\ref{fig:resnet_comparison}), indicating that this is perhaps a bigger issue with TD-learning.  Another interesting direction is to build new methods and workflows for using Q-functions that rely on the property that \methodname{} inherently represents a ``cascaded'' family of critics with different capacities---all within one network. Can this property be used for tuning network size upon deployment, cross-validation of model size, or improving efficiency of policy extraction? Answering this question would be interesting for future work. Finally, \methodname{} also provides one possibility for sequential or ``depth''-based test-time scaling for value functions. Studying how this sort of sequential scaling can be combined with parallel scaling (i.e., ensembles) and horizon reduction techniques~\citep{park2025horizon} would be interesting as well.

From a theoretical standpoint, quantifying iterative computation properties of \methodname{} would be impactful: in principle, curved flow traversals should enable the critic network to spend more test-time compute (i.e., integration steps) to perform equivalents of ``error correction'' and ``backtracking'' from large language models (LLMs)~\citep{deepseekr1_guo2025}, but now in the space of scalar, continuous values to better approximate the target Q-function. We believe formalizing this aspect would not only be impactful for value-based RL, but could also shed light on methods to use test-time compute in flow/diffusion models in other domains. Second, our results show that there are substantial representation learning benefits of \methodname{}. We believe that studying the mechanisms and differences between feature  learning induced by \methodname{} compared to standard TD-learning with regression~\citep{kumar2021dr3} or classification~\citep{farebrother2024stop} would be interesting for future value-based methods. \methodname{} also provides a rich family of auxiliary tasks to train a critic, which provides another angle to explain and study its properties. All of these are impactful directions to study in future work.

\vspace{-0.2cm}
\section*{Acknowledgements}
\vspace{-0.2cm}
We thank Max Sobol Mark, Zheyuan Hu, Aravind Venugopal, Seohong Park, Mitsuhiko Nakamoto, and Amrith Setlur for feedback on an earlier version of the paper and informative discussions. We thank members of the CMU AIRe lab for feedback and support. AK thanks Max Simchowitz for discussions on flow-matching and diffusion models. This work was supported by a Schmidt Sciences AI2050 fellowship and the Office of Naval Research under N00014-24-12206. We thank the TRC program of Google Cloud and NCSA Delta for providing computational resources that supported this work. This work used the Delta advanced computing and data resource at the National Center for Supercomputing Applications (NCSA) through allocation CIS250548 from the Advanced Cyberinfrastructure Coordination Ecosystem: Services And Support (ACCESS) program.
Delta is supported by the National Science Foundation (award OAC-2005572) and the State of Illinois, and ACCESS is supported by U.S. National Science Foundation grants 2138259, 2138286, 2138307, 2137603, and 2138296.
\bibliography{main}

\newpage
\appendix

\onecolumn 
\part*{Appendices}

\section{Additional Results for \methodname{}}

\begin{figure}[ht]
    \centering
    \includegraphics[width=0.9\textwidth, height=0.5\textwidth, keepaspectratio]{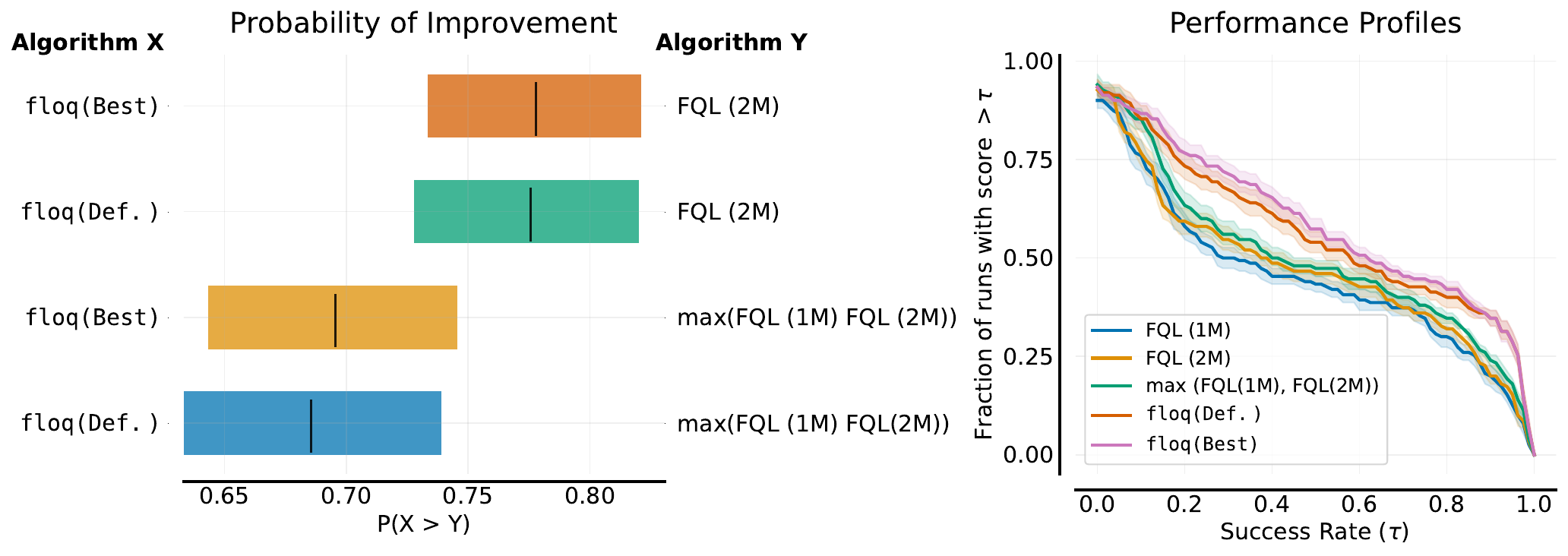}
    \caption{\footnotesize Comparison of \methodname{} against baseline FQL, following \cite{rliable_agarwal2021}. 
\textbf{Left:} Probability of Improvement $P(X > Y)$ showing that \methodname{} consistently outperform FQL across OGBench tasks. 
\textbf{Right:} Performance profiles illustrating that \methodname{} achieves higher scores across a larger fraction of runs compared to FQL.}
    \label{fig:rliable_2}
\end{figure}

\begin{figure}[ht]
\centering
\vspace{-0.2cm}

\includegraphics[width=0.99\textwidth]{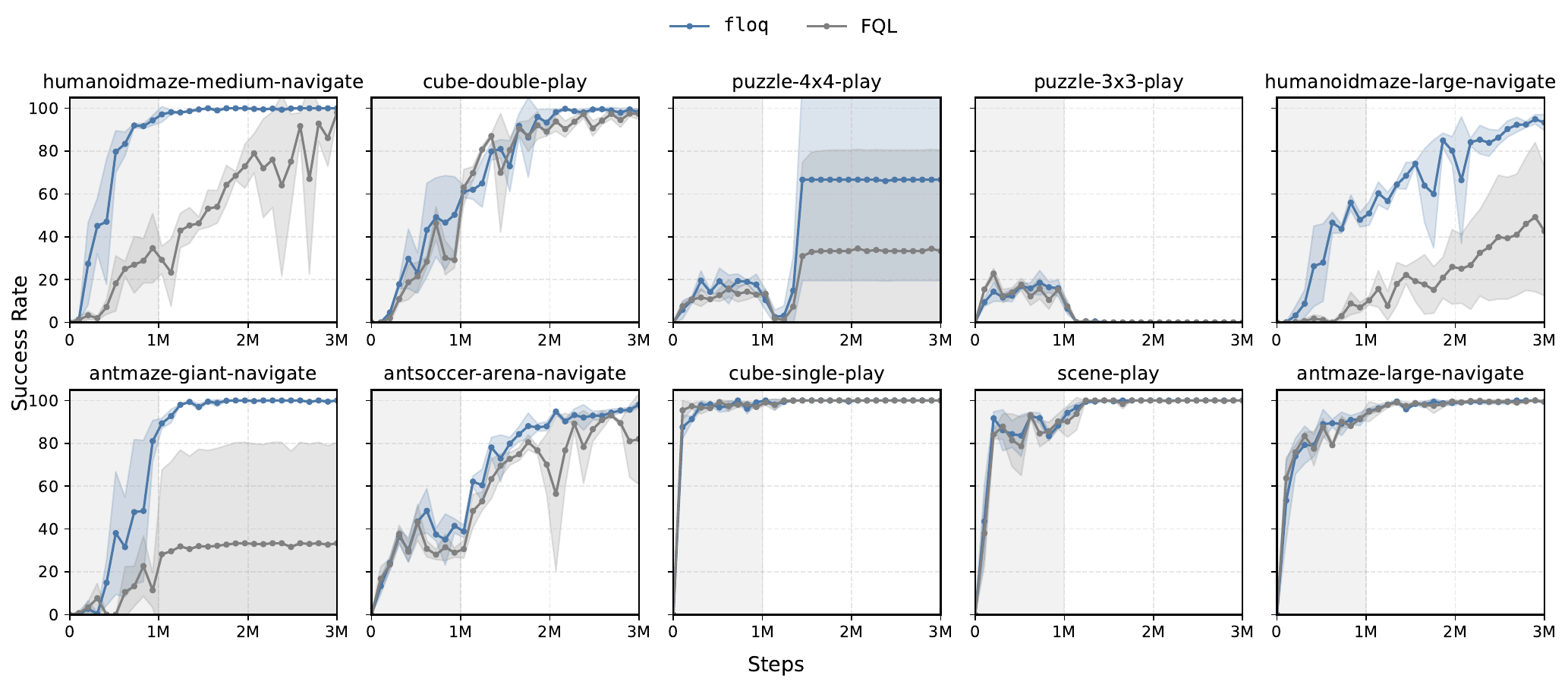}
\vspace{-0.4cm}
\caption{\footnotesize \emph{\textbf{Learning curves for online fine-tuning}} of \methodname{} and FQL across all default tasks. \methodname{} not only provides a stronger initialization from offline RL training but also maintains its advantage through online fine-tuning, leading to faster adaptation and higher final success rates. The shaded gray area denotes offline RL training.
% \nauman{imo increase the legend size in Figs 4 and 5, maybe delete the information on "4-layer MLP" to make it fit}
}
\vspace{-0.3cm}
\label{fig:online_finetuning_all}
\end{figure}

In this section, we provide some additional and complete results supplementing the ones in main paper.
\begin{enumerate}[itemsep=5pt]
    \item Figure~\ref{fig:rliable_2} presents performance profiles and $P(X>Y)$ statistic comparing \methodname{} with FQL on all the 50 tasks studied in the paper.
    \item Figure~\ref{fig:online_finetuning_all} presents results for online fine-tuning on all 10 default tasks.
\end{enumerate}

\vspace{-0.2cm}

\include{main_offline_results}

\begin{table*}[t!]
\vspace{-10pt}
\caption{
\footnotesize
\textbf{Offline RL results (Default Tasks).} \methodname{} achieves competitive or superior performance compared to the baselines. ``Hard'' tasks refers to the set of default tasks where the FQL baseline score is below 50\% performance. \methodname{} is especially more performant on these hard tasks, more than doubling FQL’s baseline performance.}
\label{table:default_task_offline_results}
\centering
\resizebox{0.99\linewidth}{!}{
\begin{threeparttable}
\begin{tabular}{l cc c cccc | cc}
\toprule
& \multicolumn{2}{c}{\texttt{Gaussian Policy}} & \multicolumn{1}{c}{\texttt{Diff.\!\!\!\! Policy}} & \multicolumn{4}{c}{\texttt{Flow Policy}} & \multicolumn{2}{c}{\texttt{Flow Q-function (Ours)}} \\
\cmidrule(lr){2-3} \cmidrule(lr){4-4} \cmidrule(lr){5-8} \cmidrule(lr){9-10}
\texttt{Env (Default Task)} & \texttt{BC} & \texttt{ReBRAC} & \texttt{DSRL} & \texttt{SORL} & \texttt{IQN} & \texttt{FQL (1M)} & \texttt{FQL(2M)} & \texttt{\methodname{} (Def.)} & \texttt{\methodname{} (Best)} \\
\midrule

\texttt{antmaze-large}   & $0$ {\tiny $\pm 0$}   & $91$ {\tiny $\pm 10$} & $40$ {\tiny $\pm 29$} & $93$ {\tiny $\pm 2$}  & $86$ {\tiny $\pm 1$}  & $80$ {\tiny $\pm 8$}  & $85$ {\tiny $\pm 4$}  & \mathbf{$94$} {\tiny $\pm 4$}  & \mathbf{$94$} {\tiny $\pm 4$} \\
\texttt{antmaze-giant}  & $0$ {\tiny $\pm 0$}   & $27$ {\tiny $\pm 22$} & $0$ {\tiny $\pm 0$}   & $12$ {\tiny $\pm 6$}  & $5$ {\tiny $\pm 6$}   & $11$ {\tiny $\pm 16$} & $14$ {\tiny $\pm 29$} & $70$ {\tiny $\pm 8$}  & \mathbf{$86$} {\tiny $\pm 4$} \\
\texttt{hmmaze-medium}  & $1$ {\tiny $\pm 0$}   & $16$ {\tiny $\pm 9$}  & $34$ {\tiny $\pm 20$} & $67$ {\tiny $\pm 4$}  & $48$ {\tiny $\pm 17$} & $19$ {\tiny $\pm 12$} & $58$ {\tiny $\pm 25$} & \mathbf{$98$} {\tiny $\pm 1$} & \mathbf{$98$} {\tiny $\pm 1$} \\
\texttt{hmmaze-large}   & $0$ {\tiny $\pm 0$}   & $2$ {\tiny $\pm 1$}   & $10$ {\tiny $\pm 12$} & $20$ {\tiny $\pm 9$}  & $35$ {\tiny $\pm 5$}  & $8$ {\tiny $\pm 5$}   & $14$ {\tiny $\pm 10$} & \mathbf{$52$} {\tiny $\pm 8$} & \mathbf{$52$} {\tiny $\pm 8$} \\
\texttt{antsoccer-arena} & $1$ {\tiny $\pm 0$}   & $0$ {\tiny $\pm 0$}   & $28$ {\tiny $\pm 0$}  & \mathbf{$54$} {\tiny $\pm 5$} & \mathbf{$54$} {\tiny $\pm 7$}  & $39$ {\tiny $\pm 6$}  & $49$ {\tiny $\pm 11$} & $49$ {\tiny $\pm 10$} & $49$ {\tiny $\pm 10$} \\
\texttt{cube-single}     & $3$ {\tiny $\pm 1$}   & $92$ {\tiny $\pm 4$}  & $93$ {\tiny $\pm 14$} & \textbf{$99$} {\tiny $\pm 0$} & $98$ {\tiny $\pm 1$}  & $96$ {\tiny $\pm 1$}  & $94$ {\tiny $\pm 5$}  & \mathbf{$99$} {\tiny $\pm 2$} & \mathbf{$99$} {\tiny $\pm 2$} \\
\texttt{cube-double}     & $0$ {\tiny $\pm 0$}   & $7$ {\tiny $\pm 3$}   & $53$ {\tiny $\pm 14$} & $33$ {\tiny $\pm 8$}  & $57$ {\tiny $\pm 2$}  & $36$ {\tiny $\pm 6$}  & $29$ {\tiny $\pm 8$}  & \mathbf{$72$} {\tiny $\pm 15$} & \mathbf{$72$} {\tiny $\pm 15$} \\
\texttt{scene}           & $1$ {\tiny $\pm 1$}   & $50$ {\tiny $\pm 13$} & $88$ {\tiny $\pm 9$}  & \mathbf{$89$} {\tiny $\pm 9$} & $80$ {\tiny $\pm 4$}  & $76$ {\tiny $\pm 9$}  & $78$ {\tiny $\pm 7$}  & $83$ {\tiny $\pm 10$} & $83$ {\tiny $\pm 10$} \\
\texttt{puzzle-3x3}      & $1$ {\tiny $\pm 1$}   & $2$ {\tiny $\pm 1$}   & $0$ {\tiny $\pm 0$}   & --                   & \mathbf{$20$} {\tiny $\pm 3$} & $16$ {\tiny $\pm 5$}  & $14$ {\tiny $\pm 4$}  & $17$ {\tiny $\pm 6$}  & $17$ {\tiny $\pm 6$} \\
\texttt{puzzle-4x4}      & $0$ {\tiny $\pm 0$}   & $10$ {\tiny $\pm 3$}  & \mathbf{$37$} {\tiny $\pm 13$} & --                   & $16$ {\tiny $\pm 1$}  & $11$ {\tiny $\pm 3$}  & $5$ {\tiny $\pm 2$}   & $12$ {\tiny $\pm 4$}  & $19$ {\tiny $\pm 5$} \\

\midrule
\texttt{Average Score (All Tasks)}  & $1$  & $30$ & $38$ & -- &  $49$ & $40$ & $44$ & $64$ & \mathbf{$66$}\\
\texttt{Average Score (Hard Tasks)} & $0$  & $8$  & $21$ & -- & $30$ & $20$ & $21$ & $45$ & \mathbf{$50$} \\
\bottomrule
\end{tabular}
\end{threeparttable}}
\end{table*}
\setlength{\tabcolsep}{3pt}
\begin{table}[t!]
\caption{\footnotesize \textbf{\emph{FQL ResNet performace on \texttt{humanoidmaze-large} and \texttt{antmaze-giant}}}. 
$(m,n)$ indicates $n$ blocks each of depth $m$. For each fixed number of \texttt{FLOPs} $m \times n$, 
the best-performing architecture per environment is in bold.}
\centering
\footnotesize
\begin{tabular}{c|c|c}
\toprule
\texttt{FLOPs} $(m \times n)$ & \texttt{HM-Large} & \texttt{Antmaze-Giant} \\
\midrule
$4$   & $\mathbf{35}$ {\scriptsize $\pm$ 19} (2,2) & $\mathbf{13}$ {\scriptsize $\pm$ 12} (2,2) \\
      & $14$ {\scriptsize $\pm$ 10} (4,1) & $11$ {\scriptsize $\pm$ 16} (4,1) \\
\midrule
$8$   & $\mathbf{46}$ {\scriptsize $\pm$ 11} (2,4) & $31$ {\scriptsize $\pm$ 13} (2,4) \\
      & $21$ {\scriptsize $\pm$ 11} (4,2) & $22$ {\scriptsize $\pm$ 9} (4,2) \\
      & $22$ {\scriptsize $\pm$ 11} (8,1) & $\mathbf{32}$ {\scriptsize $\pm$ 14} (8,1) \\
\midrule
$16$  & $\mathbf{41}$ {\scriptsize $\pm$ 7} (2,8) & $32$ {\scriptsize $\pm$ 13} (2,8) \\
      & $34$ {\scriptsize $\pm$ 8} (4,4)  & $17$ {\scriptsize $\pm$ 8} (4,4) \\
      & $26$ {\scriptsize $\pm$ 9} (8,2) & $\mathbf{46}$ {\scriptsize $\pm$ 11} (8,2) \\
      & $24$ {\scriptsize $\pm$ 22} (16,1)  & $0$ {\scriptsize $\pm$ 0} (16,1) \\
\midrule
$32$  & $25$ {\scriptsize $\pm$ 10} (2,16)  & $23$ {\scriptsize $\pm$ 11} (2,16) \\
      & $28$ {\scriptsize $\pm$ 13} (4,8) & $18$ {\scriptsize $\pm$ 14} (4,8) \\
      & $\mathbf{38}$ {\scriptsize $\pm$ 19} (8,4) & $\mathbf{30}$ {\scriptsize $\pm$ 9} (8,4) \\
      & $0$ {\scriptsize $\pm$ 0} (16,2) & $0$ {\scriptsize $\pm$ 0} (16,2) \\
\bottomrule
\end{tabular}
\end{table}

\vspace{-0.2cm}
\section{Hyperparameters and Additional Details}
\label{app:details}
\vspace{-0.2cm}
In this section, we present some details for \methodname{} that we could not cover in the main paper, along with a pseudocode and a complete list of hyperparameters used by our approach.

\begin{algorithm}[t]
\caption{Critic Flow Matching (\methodname) in conjunction with FQL~\citep{park2025flow}}
\label{alg:cfm}
\begin{algorithmic}
\footnotesize
\State \textbf{Given:} offline dataset of transitions $\mathcal{D}$, 
\State \textbf{Models:} a flow critic, $Q^\text{FLOW}_\theta(s, a, \bz)$, a distilled critic, $Q^\text{distilled}_\psi(s, a, \bz)$, a flow policy $\pi_\phi(\cdot|\bs)$, one-step policy $\mu_\omega(s, \cdot)$.
\BeginBox[fill=myblue!8]
\Function{$Q^{\text{flow}}_\theta(\bs, \ba, \bz)$}{} \Comment{\color{myblue} Flow Q-function, introduced by \methodname{}}
\For{$t = 0, 1, \dots, K-1$}
\State $\bz(t+1) \gets \bz(t) + \nicefrac{1}{K} \cdot v_\theta\left(\nicefrac{t}{K}, \bz(t) ~|~ s,a\right)$ \Comment{Euler method, time $t$ is normalized}
\EndFor
\State \Return $\bz(K)$
\EndFunction
\EndBox

\BeginBox[fill=white]
\Function{$\pi_{\phi}(\ba | \bs)$}{} \Comment{\color{myblue} Flow policy from FQL, though policy training is orthogonal to \methodname{}}
\For{$t = 0, 1, \dots, M-1$}
\State Sample $\bx(0) \sim \mathcal{N}(0, I_d)$
\State $\bx(t+1) \gets \bx(t) + \nicefrac{1}{M} \cdot w_\phi\left(\nicefrac{t}{M}, \bx(t) ~|~ \bs \right)$ \Comment{Euler method, time t is normalized}
\EndFor
\State \Return $\bx(M)$
\EndFunction
\EndBox
\vspace{5pt}

\While{not converged}

\State Sample batch $\{(s, a, r, s')\} \sim \gD$

\BeginBox[fill=myblue!8]
\LComment{\color{myblue} Train vector field $v_\theta$ in flow critic $Q^\text{FLOW}_\theta$}
\State $a' \leftarrow \mathrm{Sample}(\pi_\phi(\cdot|\bs'))$ \Comment{Sample actions from policy, typically the one-step policy for FQL}
\State $\bz(0) \sim \mathrm{Unif}~[l,u], \bz'(0)_{1:m} \sim \mathrm{Unif}~[l,u]$ \Comment{Sample initial noise for computing the Q-value}
\State $\bz(1) \gets r + \gamma \cdot \nicefrac{1}{m} \cdot \sum_{i=1}^m Q^{\text{FLOW}}_{\bar{\theta}}(s', a', \bz'_i(0))$  \Comment{Use noise $\bz'_i(0)$ for computing TD-target}
\State $\bz(t) \gets (1-t) \cdot \bz(0) + t \cdot \bz(1)$ \Comment{Compute interpolant $\bz(t)$ for random $t$}
\State Update {\color{myblue}$\theta$} to minimize $\E \left[(v_{\color{myblue}\theta}\left(t, \bz(t) ~|~ s, a\right) - \big(\bz(1) - \bz(0) \big))^2 \right]$ \Comment{Linear flow-matching loss}
\EndBox
\BeginBox[fill=myblue!8]
\LComment{\color{myblue} Train distill critic $Q^{\text{distill}}_\psi$ for policy extraction}
\State Update {\color{myblue}$\psi$} to minimize $\E_{\bz(0)} \left[(Q^{\text{distill}}_{\psi}(s,a) - Q^{\text{FLOW}}_{\theta}(s,a, \bz(0)))^2 \right]$
\EndBox

\BeginBox[fill=white]
\LComment{\color{myblue} Train a BC flow policy $\pi_\phi$, analogous to FQL}
\State $\bx(0) \sim \gN(0, I_d)$
\State $\bx(1) \gets a$
\State $t \sim \mathrm{Unif}([0, 1])$
\State $\bx(t) \gets (1-t) \cdot \bx(0) + t \cdot \bx(1)$ \Comment{For FQL policy, compute policy interpolant}
\State Update {\color{myblue}$\phi$} to minimize $\E \left[\|w_{\color{myblue}\phi}\left(t, \bx(t) | s) - (\bx(1) - \bx(0)\right)\||_2^2\right]$ \Comment{Flow-matching loss for policy}
\EndBox

\BeginBox[fill=white]
\LComment{\color{myblue} Train one-step policy $\mu_\omega$ to maximize the learned distill critic while staying close to BC flow policy}
\State $\bx \sim \gN(0, I_d)$
\State $a^\pi \gets \mu_{\color{myblue}\omega}(s, \bx)$
\State Update {\color{myblue}$\omega$} to minimize $\E \left[-Q^{\text{distill}}_\psi(s, a^\pi) + \alpha \|a^\pi - \pi_\phi(s, z)\|_2^2 \right]$
\EndBox
\EndWhile
\Return One-step policy $\pi_\omega$
\end{algorithmic}
\end{algorithm}

\textbf{Efficient policy extraction using a distilled critic.} 
Because \methodname{} parameterizes a flow-matching critic, extracting reparameterized policy gradients requires computing gradients of the full integration process with respect to the input action, which is a costly operation especially when using many integration steps. To reduce this overhead, we adapt a technique introduced by \citet{park2025flow} for flow-matching policies and apply it to critics. Specifically, we train a \emph{distilled critic}, $Q^\mathrm{distill}_\psi(s,a)$, to approximate the predictions obtained by integrating the flow critic, $Q^\mathrm{flow}_\theta(s,a,\bz)$. Policy extraction is then performed directly on the distilled critic. Importantly, the distilled critic is not conditioned on the noise $\bz$, since in practice we only use the mean prediction of the flow critic. This design allows the distilled critic to implicitly capture that behavior while eliminating unnecessary conditioning. We illustrate this idea in Algorithm~\ref{alg:cfm}.

\textbf{Hyperparameters for offline RL results.} Following Park et al. \cite{park2025flow}, we tune the BC coefficient $\alpha$ on the \texttt{default-task} of each environment and then fix this value for the remaining tasks. For both FQL and \methodname{}, $\alpha$ is tuned over $\{\alpha_{\mathrm{FQL}} - \Delta,\ \alpha_{\mathrm{FQL}},\ \alpha_{\mathrm{FQL}} + \Delta\}$, where $\Delta = 100$ for the \texttt{puzzle}, \texttt{cube}, and \texttt{scene} environments, and $\Delta = 10$ for the \texttt{ant} and \texttt{humanoid} environments. The baseline values $\alpha_{\mathrm{FQL}}$ are taken from Table 6 in Park et al. \cite{park2025flow}, and the final values for both methods are reported in Table \ref{table:cfm_task_specific_alpha}. For \methodname{}, after tuning $\alpha$ with the default configuration ($K = 8$ flow steps and width $(u-l) = \kappa (Q_{\max} - Q_{\min})$ with $\kappa = 0.1$), we tune $K \in \{4, 8\}$ and $\kappa \in \{0.1, 0.25\}$ on the \texttt{default-task} of each environment. These values, referred to as \methodname{(Best)}, are reported in Table \ref{table:cfm_task_specific_hyperparameters}.  In all cases, for \methodname{}, we utilize $m=8$ samples of initial noise to compute the target Q-value as discussed in Section~\ref{sec:training}.

\textbf{Hyperparameters for online fine-tuning.} Most hyperparameters (unless otherwise stated) remain similar in online fine-tuning and offline RL pre-training.  For both FQL and \methodname{}, $\alpha$ is tuned in the range $[10,100]$ (step size 10) for the \texttt{ant} and \texttt{humanoid} environments, and in $[100,1000]$ (step size 100) for the \texttt{cube}, \texttt{scene}, and \texttt{puzzle} environments. The selected $\alpha$ values are given in Table \ref{table:cfm_task_specific_alpha_online_ft}.

For \methodname{}, after tuning $\alpha$ with the default configuration ($K = 8$, $\kappa = 0.1$), we tune $K \in \{4, 8, 16\}$ and $\kappa \in \{0.1, 0.25\}$ per environment. The chosen values are reported in Table \ref{table:cfm_task_specific_online_ft_hyperparameters}.

\textbf{Number of seeds.} We ran $3$ seeds for each configuration of both \texttt{floq} and \textbf{FQL} on each task, for both offline RL and online fine-tuning.

\begin{table}[t!]
\centering
\caption{\footnotesize{\textbf{Hyperparameters for \methodname{}.} Differences from FQL are shown in \textcolor{blue!50}{light blue} within brackets. Other hyperparameters are kept to be the same as FQL.}}
\label{table:cfm_fql_hyperparameters}
\resizebox{0.99\linewidth}{!}{%
\begin{tabular}{lc}
\toprule
\textbf{Hyperparameter} & \textbf{Value (\methodname{})} \\
\midrule
Learning rate & 0.0003 \\
Optimizer & Adam (Kingma \& Ba, 2015 \cite{adam_kingma2015}) \\
Gradient steps & 2M (Offline), 1M + 2M (Online FT) \\
Minibatch size & 256 (default), 512 for \texttt{hm-large}, \texttt{antmaze-giant} \\
Flow $Q$ Network MLP dims & [512,512,512,512] (default), [512,512] for \texttt{cube} envs \\
Distill $Q$ MLP dims & [512,512,512,512] \textcolor{blue!50}{\footnotesize (not used in FQL)} \\
Nonlinearity & GELU (Hendrycks \& Gimpel, 2016 \cite{gelu_hendrycks2016}) \\
Target network smoothing coeff. & 0.005 \\
Discount factor $\gamma$ & 0.99 (default), 0.995 for \texttt{antmaze-giant}, \texttt{humanoidmaze}, \texttt{antsoccer} \\
Flow time sampling distribution & $\text{Unif}([0, 1])$ \\
Clipped double Q-learning & False (default), True (\texttt{antmaze-giant}) \textcolor{blue!50}{\footnotesize (+ \texttt{antmaze-large} in FQL)} \\
BC coefficient $\alpha$ & Tables \ref{table:cfm_task_specific_alpha}, \ref{table:cfm_task_specific_alpha_online_ft} \\
Actor Flow steps & 10 \\
Critic Flow steps & 8 (default), Table \ref{table:cfm_task_specific_hyperparameters} for env-wise \textcolor{blue!50}{\footnotesize (not used in FQL)} \\
Initial Sample Range & 0.1 (default), Table \ref{table:cfm_task_specific_hyperparameters} for env-wise \textcolor{blue!50}{\footnotesize (not used in FQL)} \\
Number Of Initial Noise Samples & 8 \textcolor{blue!50}{\footnotesize (not used in FQL)} \\
Fourier Time Embed Dimension & 64 \textcolor{blue!50}{\footnotesize (not used in FQL)} \\
\bottomrule
\end{tabular}}
\end{table}

\begin{table}[t!]
\centering
\caption{\footnotesize{Environment-wise BC-Coefficient ($\alpha$) for FQL and \methodname{} (Offline RL).}}
\label{table:cfm_task_specific_alpha}
\begin{tabular}{lc}
\toprule
\textbf{\texttt{Environment ($5$ tasks each)}} & \textbf{$\alpha$ (FQL), $\alpha$ (\methodname{})} \\
\midrule
\texttt{antmaze-large} & $10, 10$\\ 
\texttt{antmaze-giant} & $10, 10$\\ 
\texttt{hmmaze-medium} & $30, 30$\\ 
\texttt{hmmaze-large} & $30, 20$\\  
\texttt{antsoccer-arena} & $10, 10$\\ 
\texttt{cube-single} & $300, 300$\\
\texttt{cube-double} & $300, 300$\\
\texttt{scene-play} & $300, 300$\\
\texttt{puzzle-3x3} & $1000, 1000$\\
\texttt{puzzle-4x4} & $1000, 1000$\\
\bottomrule
\end{tabular}
\end{table}
\begin{table}[t!]
\centering
\caption{\footnotesize{Environment-wise BC-Coefficient ($\alpha$) for FQL and \methodname{} (Online Fine-Tuning).}}
\label{table:cfm_task_specific_alpha_online_ft}
\begin{tabular}{lc}
\toprule
\textbf{\texttt{Environment ($5$ tasks each)}} & \textbf{$\alpha$ (FQL), $\alpha$ (\methodname{})} \\
\midrule
\texttt{antmaze-large} & $10, 10$\\ 
\texttt{antmaze-giant} & $10, 10$\\ 
\texttt{hmmaze-medium} & $80, 30$\\ 
\texttt{hmmaze-large} & $40, 20$\\  
\texttt{antsoccer-arena} & $30, 30$\\ 
\texttt{cube-single} & $300, 300$\\
\texttt{cube-double} & $300, 300$\\
\texttt{scene-play} & $300, 300$\\
\texttt{puzzle-3x3} & $1000, 1000$\\
\texttt{puzzle-4x4} & $1000, 1000$\\
\bottomrule
\end{tabular}
\end{table}

\begin{table}[t!]
\centering
\caption{\footnotesize{Environment-wise Initial Sample Range ($\frac{u-l}{Q_{\max} - Q_{\min}}$) and Flow Steps ($K$) for \methodname{} (Best)} (Offline RL).}
\label{table:cfm_task_specific_hyperparameters}
\begin{tabular}{lc}
\toprule
\textbf{\texttt{Environment ($5$ tasks each)}} & (\textbf{$\frac{u-l}{Q_{\max} - Q_{\min}}$, $K$)} \\
\midrule
\texttt{antmaze-large} & $(0.1, 8)$\\ 
\texttt{antmaze-giant} & $(0.1, 4)$\\ 
\texttt{hmmaze-medium} & $(0.1, 8)$\\ 
\texttt{hmmaze-large} &  $(0.1, 8)$\\  
\texttt{antsoccer-arena} & $(0.1, 8)$\\ 
\texttt{cube-single} & $(0.1, 8)$\\
\texttt{cube-double} & $(0.1, 8)$\\
\texttt{scene-play} & $(0.1, 8)$\\
\texttt{puzzle-3x3} & $(0.1, 8)$\\
\texttt{puzzle-4x4} & $(0.25, 8)$\\
\bottomrule
\end{tabular}
\end{table}

\begin{table}[t!]
\centering
\caption{\footnotesize{Environment-wise Initial Sample Range ($\frac{u-l}{Q_{\max} - Q_{\min}}$) and Flow Steps ($K$) for \methodname{}} (Online FT).}
\label{table:cfm_task_specific_online_ft_hyperparameters}
\begin{tabular}{lc}
\toprule
\textbf{\texttt{Environment ($5$ tasks each)}} & (\textbf{$\frac{u-l}{Q_{\max} - Q_{\min}}$, $K$)} \\
\midrule
\texttt{antmaze-large} & $(0.1, 8)$\\ 
\texttt{antmaze-giant} & $(0.1, 4)$\\ 
\texttt{hmmaze-medium} & $(0.1, 8)$\\ 
\texttt{hmmaze-large} &  $(0.1, 16)$\\  
\texttt{antsoccer-arena} & $(0.1, 8)$\\ 
\texttt{cube-single} & $(0.1, 8)$\\
\texttt{cube-double} & $(0.1, 8)$\\
\texttt{scene-play} & $(0.1, 8)$\\
\texttt{puzzle-3x3} & $(0.1, 8)$\\
\texttt{puzzle-4x4} & $(0.25, 8)$\\
\bottomrule
\end{tabular}
\end{table}

\section{Wall Clock Run-Time}
We report the wall clock run-times for FQL and  \methodname{} in Table \ref{tab:runtime}. 
\begin{table}[ht]
\centering
\footnotesize
\caption{Total wall-clock runtime (in $10^3$ seconds) for FQL and \methodname{} with varying numbers of flow integration steps across four representative environments. Reported numbers correspond to 2M training steps.}
\begin{tabular}{l c cccccc}
\toprule
\multirow{2}{*}{Environment} & \multirow{2}{*}{FQL} & \multicolumn{5}{c}{\methodname{} (Flow Steps)} \\
\cmidrule(lr){3-7}
& & 1 & 2 & 4 & 8 & 16 \\
\midrule
HM-Maze Large   & 14 & 24 & 28 & 35 & 50 & 79 \\
HM-Maze Medium  & 12 & 19 & 21 & 23 & 30 & 47 \\
Cube-Double     & 10 & 15 & 16 & 17 & 19 & 26 \\
Antmaze-Giant   & 10 & 20 & 24 & 30 & 45 & 74 \\
\bottomrule
\end{tabular}
\label{tab:runtime}
\end{table}
\vspace{-0.2cm}
\section{Flow Visualizations}
\begin{figure}[t]
\centering
\vspace{-0.2cm}
\includegraphics[width=0.99\textwidth]{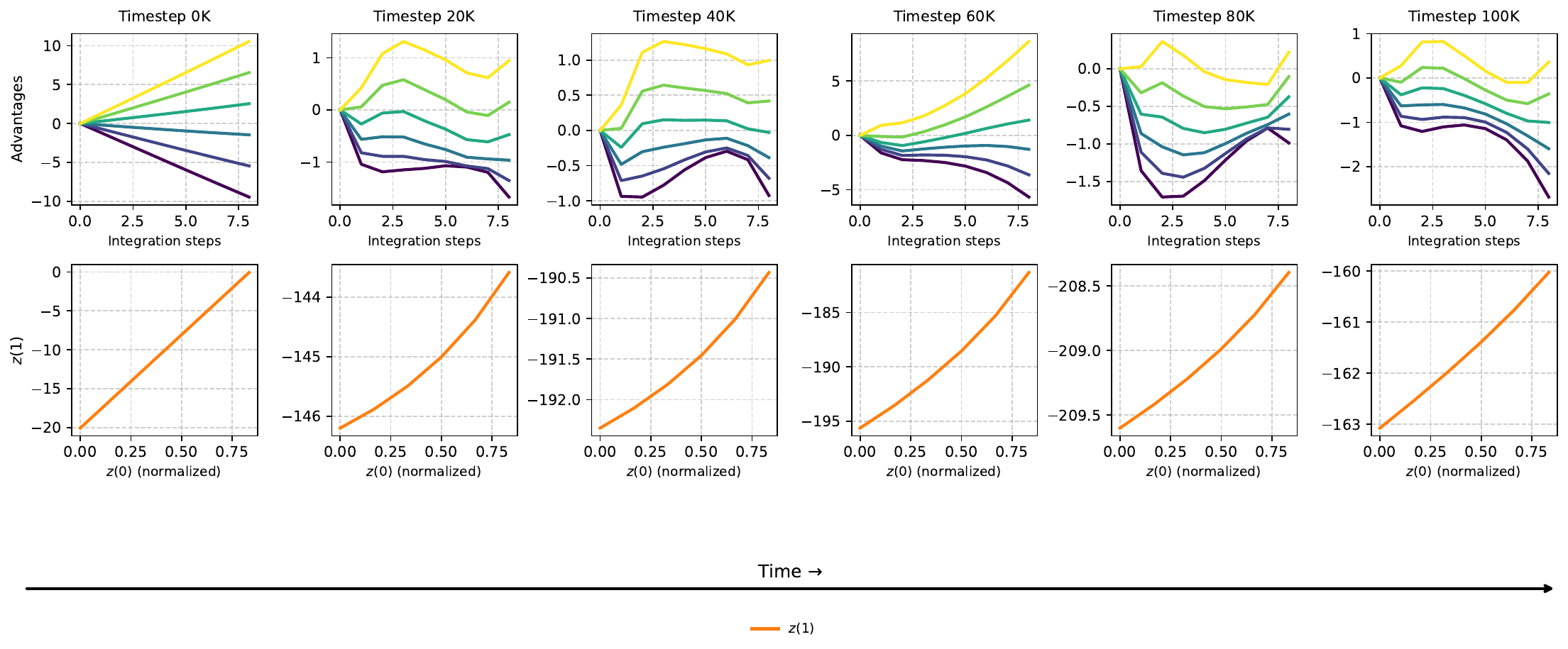}
\vspace{-0.3cm}
\caption{\footnotesize 
\emph{\textbf{Visualizing the evolution of the trajectories of the flow critic during training.}} 
}
\label{fig:flow_evolution}
\vspace{-0.4cm}
\end{figure}

\vspace{-0.2cm}
We visualize the evolution of the learned flow critic during training on \texttt{cube-double} in Figure~\ref{fig:flow_evolution}, with $\kappa = 0.1$. Because raw Q-values can have large magnitudes, directly plotting them makes it difficult to assess the curvature of the learned flow. Instead, we plot advantage values, defined as the gap between the predicted Q-value obtained by integrating for $k$ flow steps at various noise samples $\bz_i(0)$, namely $\psi(k, \bz_i(k) \mid s,a)$ for $i \in [5], , k \in [1, \ldots, K]$, and the expected value of that state–action pair after $K$ steps, scaled linearly to $k$ steps. Put simply, this advantage quantifies how far the intermediate estimate $\psi(k, \bz_i(k) \mid s,a)$ deviates from the ``straight line'' path between the initial noise sample $\bz_i(0)$ and the final Q-value. We find that these deviations are consistently non-zero and vary substantially across the integration process. In many cases, they exhibit a characteristic pattern of overshooting followed by correction: larger deviations early on that diminish as integration proceeds. These dynamics provide direct evidence that the learned flows follow curved rather than linear trajectories. We also visualize the final Q-value output $\bz(1)$ as a function of the input $\bz(0)$ in Figure~\ref{fig:flow_evolution} (bottom) and find that the final $\bz(1)$ depends non-linearly on the initial noise value.

\end{document}

%% file: math_commands.tex
\usepackage{amsmath,amsfonts,bm}

\def\eqref#1{equation~\ref{#1}}
\def\1{\bm{1}}

\DeclareMathAlphabet{\mathsfit}{\encodingdefault}{\sfdefault}{m}{sl}
\SetMathAlphabet{\mathsfit}{bold}{\encodingdefault}{\sfdefault}{bx}{n}

\def\gD{{\mathcal{D}}}

\def\gN{{\mathcal{N}}}

\newcommand{\E}{\mathbb{E}}

\newcommand{\methodname}{\texttt{{floq}}}

\newcommand{\behavior}{{\pi_\beta}}

\newcommand{\bx}{\mathbf{x}}
\newcommand{\bz}{\mathbf{z}}

\newcommand{\bs}{s}
\newcommand{\ba}{a}

\renewcommand{\mathbf}{\boldsymbol}

\newcommand{\mc}{\mathcal}

\newcommand{\bb}{\mathbb}

\makeatletter
\def\Ddots{\mathinner{\mkern1mu\raise\p@
\vbox{\kern7\p@\hbox{.}}\mkern2mu
\raise4\p@\hbox{.}\mkern2mu\raise7\p@\hbox{.}\mkern1mu}}
\makeatother
\newcommand{\brac}[1]{\left[ #1 \right]}

\numberwithin{equation}{section}

%% file: main_offline_results.tex
% DEFAULT TASK PERF

\setlength{\tabcolsep}{3pt}
\begin{table*}[t!]
\vspace{-10pt}

\caption{
\footnotesize
\textbf{Offline RL results (all tasks).} \methodname{} performs comparable or superior to the baselines on most tasks. \texttt{(*)} denotes the default task per environment, and this choice is consistent per the selections made by \citet{park2025flow}. SORL~\citep{espinosa2025scaling} did not report results on two of the domains, hence we report their numbers as ``--''.}

\label{table:all_task_offline_results}
\centering
\vspace{-0.1cm}
\resizebox{0.93\linewidth}{!}
{
\begin{threeparttable}
\begin{tabular}{l cc ccc | cc}

\toprule

\multicolumn{1}{c}{} & \multicolumn{2}{c}{\texttt{Gaussian Policy}} & \multicolumn{3}{c}{\texttt{Flow Policy}} & \multicolumn{2}{c}{\texttt{Flow Q-function (Ours)}} \\
\cmidrule(lr){2-3} \cmidrule(lr){4-6} \cmidrule(lr){7-8}
\texttt{Environment (5 tasks each)} & \texttt{BC} & \texttt{ReBRAC} & \texttt{SORL} &\texttt{FQL (1M)} & \texttt{FQL(2M)} &\texttt{\methodname{} (Def.)} &\texttt{\methodname{} (Best)} \\

\midrule
\texttt{antmaze-large-task1(*)}  &$0$ {\tiny $\pm 0$}& $91$ {\tiny $\pm 10$}  & $93$ {\tiny $\pm 2$} & $80$ {\tiny $\pm 8$}& $85$ {\tiny $\pm 4$}& \mathbf{$94$} {\tiny $\pm 4$} & \mathbf{$94$} {\tiny $\pm 4$}\\
\texttt{antmaze-large-task2}  &$6$ {\tiny $\pm 3$}& \mathbf{$88$} {\tiny $\pm 4$}  & $79$ {\tiny $\pm 5$} & $57$ {\tiny $\pm 10$}& $64$ {\tiny $\pm 7$}& $82$ {\tiny $\pm 5$} & $82$ {\tiny $\pm 5$}\\
\texttt{antmaze-large-task3}  &$29$ {\tiny $\pm 5$}& $51$ {\tiny $\pm 18$}  & $88$ {\tiny $\pm 10$} & $93$ {\tiny $\pm 3$}& $95$ {\tiny $\pm 2$}& \mathbf{$96$} {\tiny $\pm 4$} & \mathbf{$96$} {\tiny $\pm 4$}\\
\texttt{antmaze-large-task4}  &$8$ {\tiny $\pm 3$}& $84$ {\tiny $\pm 7$}  & $91$ {\tiny $\pm 2$} & $80$ {\tiny $\pm 4$}& $86$ {\tiny $\pm 5$}& \mathbf{$90$} {\tiny $\pm 8$} & \mathbf{$90$} {\tiny $\pm 8$}\\
\texttt{antmaze-large-task5}  &$10$ {\tiny $\pm 3$}& $90$ {\tiny $\pm 2$}  & \mathbf{$95$} {\tiny $\pm 0$} & $83$ {\tiny $\pm 4$}& $87$ {\tiny $\pm 5$}& \mathbf{$95$} {\tiny $\pm 4$} & \mathbf{$95$} {\tiny $\pm 4$}\\
\midrule
\texttt{antmaze-giant-task1(*)} \textcolor{pink}{\textit{(Hard)}} &$0$ {\tiny $\pm 0$} & $27$ {\tiny $\pm 22$}  & $12$ {\tiny $\pm 6$} & $14$ {\tiny $\pm 19$} & $11$ {\tiny $\pm 16$}& $70$ {\tiny $\pm 8$} & \mathbf{$86$} {\tiny $\pm 4$}\\
\texttt{antmaze-giant-task2} \textcolor{pink}{\textit{(Hard)}} &$0$ {\tiny $\pm 0$} & $16$ {\tiny $\pm 17$}  & $0$ {\tiny $\pm 0$} & $58$ {\tiny $\pm 17$} & \mathbf{$68$} {\tiny $\pm 17$}& $59$ {\tiny $\pm 24$} & $66$ {\tiny $\pm 11$}\\
\texttt{antmaze-giant-task3} \textcolor{pink}{\textit{(Hard)}} &$0$ {\tiny $\pm 0$} & $34$ {\tiny $\pm 22$}  & $0$ {\tiny $\pm 0$} & $0$ {\tiny $\pm 0$} & $3$ {\tiny $\pm 4$}& \mathbf{$4$} {\tiny $\pm 7$} & $0$ {\tiny $\pm 0$}\\
\texttt{antmaze-giant-task4} \textcolor{pink}{\textit{(Hard)}} &$0$ {\tiny $\pm 0$} & $5$ {\tiny $\pm 12$}  & $25$ {\tiny $\pm 18$} & \mathbf{$32$} {\tiny $\pm 33$} & $31$ {\tiny $\pm 36$}& $13$ {\tiny $\pm 8$} & $17$ {\tiny $\pm 23$}\\
\texttt{antmaze-giant-task5} \textcolor{pink}{\textit{(Hard)}} &$1$ {\tiny $\pm 1$} & $49$ {\tiny $\pm 22$}  & $6$ {\tiny $\pm 15$} & $4$ {\tiny $\pm 7$} & $22$ {\tiny $\pm 31$}& $34$ {\tiny $\pm 38$} & \mathbf{$84$} {\tiny $\pm 8$}\\
\midrule
\texttt{hmmaze-medium-task1(*)}  &$1$ {\tiny $\pm 0$}& $16$ {\tiny $\pm 9$}  & $67$ {\tiny $\pm 4$} & $19$ {\tiny $\pm 12$}& $58$ {\tiny $\pm 25$}& \mathbf{$98$} {\tiny $\pm 1$} & \mathbf{$98$} {\tiny $\pm 1$}\\
\texttt{hmmaze-medium-task2}  &$1$ {\tiny $\pm 0$}& $18$ {\tiny $\pm 16$}  & $89$ {\tiny $\pm 3$} & $94$ {\tiny $\pm 3$}& $96$ {\tiny $\pm 3$}& \mathbf{$98$} {\tiny $\pm 2$} & \mathbf{$98$} {\tiny $\pm 2$}\\
\texttt{hmmaze-medium-task3}  &$6$ {\tiny $\pm 2$}& $36$ {\tiny $\pm 13$}  & $83$ {\tiny $\pm 4$} & $74$ {\tiny $\pm 18$}& $75$ {\tiny $\pm 28$}& \mathbf{$99$} {\tiny $\pm 2$} & \mathbf{$99$} {\tiny $\pm 2$}\\
\texttt{hmmaze-medium-task4}  &$0$ {\tiny $\pm 0$}& $15$ {\tiny $\pm 16$}  & $1$ {\tiny $\pm 0$} & $3$ {\tiny $\pm 4$}& \mathbf{$16$} {\tiny $\pm 23$}& \mathbf{$16$} {\tiny $\pm 23$} & \mathbf{$16$} {\tiny $\pm 23$}\\
\texttt{hmmaze-medium-task5}  &$2$ {\tiny $\pm 1$}& $24$ {\tiny $\pm 20$}  & $81$ {\tiny $\pm 20$} & $97$ {\tiny $\pm 2$}& \mathbf{$99$} {\tiny $\pm 1$}& \mathbf{$99$} {\tiny $\pm 2$} & \mathbf{$99$} {\tiny $\pm 2$}\\
\midrule
\texttt{hmmaze-large-task1(*)} \textcolor{pink}{\textit{(Hard)}} &$0$ {\tiny $\pm 0$}& $2$ {\tiny $\pm 1$}  & $20$ {\tiny $\pm 9$} & $8$ {\tiny $\pm 5$}& $14$ {\tiny $\pm 10$}& \mathbf{$52$} {\tiny $\pm 8$} & \mathbf{$52$} {\tiny $\pm 8$}\\
\texttt{hmmaze-large-task2} \textcolor{pink}{\textit{(Hard)}} &$0$ {\tiny $\pm 0$}& $0$ {\tiny $\pm 0$}  & $0$ {\tiny $\pm 0$} & $0$ {\tiny $\pm 0$}& $0$ {\tiny $\pm 0$}& \mathbf{$0$} {\tiny $\pm 1$} & \mathbf{$0$} {\tiny $\pm 1$}\\
\texttt{hmmaze-large-task3} \textcolor{pink}{\textit{(Hard)}} &$1$ {\tiny $\pm 1$}& $8$ {\tiny $\pm 4$}  & $5$ {\tiny $\pm 2$} & $10$ {\tiny $\pm 8$}& $18$ {\tiny $\pm 4$}& \mathbf{$28$} {\tiny $\pm 12$} & \mathbf{$28$} {\tiny $\pm 12$}\\
\texttt{hmmaze-large-task4} \textcolor{pink}{\textit{(Hard)}} &$1$ {\tiny $\pm 0$}& $1$ {\tiny $\pm 1$}  & $0$ {\tiny $\pm 0$} & $15$ {\tiny $\pm 9$}& \mathbf{$24$} {\tiny $\pm 7$}& $22$ {\tiny $\pm 12$} & $22$ {\tiny $\pm 12$}\\
\texttt{hmmaze-large-task5} \textcolor{pink}{\textit{(Hard)}} &$0$ {\tiny $\pm 1$}& $2$ {\tiny $\pm 2$}  & $0$ {\tiny $\pm 0$} & $12$ {\tiny $\pm 7$}& $22$ {\tiny $\pm 9$}& \mathbf{$38$} {\tiny $\pm 9$} & \mathbf{$38$} {\tiny $\pm 9$}\\
\midrule
\texttt{antsoccer-arena-task1}  &$2$ {\tiny $\pm 1$} & $0$ {\tiny $\pm 0$}  & \mathbf{$93$} {\tiny $\pm 4$} & $77$ {\tiny $\pm 4$}& $78$ {\tiny $\pm 7$}& $92$ {\tiny $\pm 4$} & $92$ {\tiny $\pm 4$}\\
\texttt{antsoccer-arena-task2}  &$2$ {\tiny $\pm 2$} & $0$ {\tiny $\pm 1$}  & \mathbf$96$ {\tiny $\pm 2$} & $88$ {\tiny $\pm 3$}& $89$ {\tiny $\pm 4$}& \mathbf{$97$} {\tiny $\pm 2$} & \mathbf{$97$} {\tiny $\pm 2$}\\
\texttt{antsoccer-arena-task3}  &$0$ {\tiny $\pm 0$} & $0$ {\tiny $\pm 0$}  & \mathbf{$55$} {\tiny $\pm 6$} & $53$ {\tiny $\pm 18$}& $61$ {\tiny $\pm 6$}& $58$ {\tiny $\pm 11$} & $58$ {\tiny $\pm 11$}\\
\texttt{antsoccer-arena-task4(*)}  &$1$ {\tiny $\pm 0$} & $0$ {\tiny $\pm 0$}  & \mathbf{$54$} {\tiny $\pm 5$} & $39$ {\tiny $\pm 6$}& $49$ {\tiny $\pm 11$}& $49$ {\tiny $\pm 10$} & $49$ {\tiny $\pm 10$}\\
\texttt{antsoccer-arena-task5}  &$0$ {\tiny $\pm 0$} & $0$ {\tiny $\pm 0$}  & \mathbf{$47$} {\tiny $\pm 9$} & $36$ {\tiny $\pm 9$}& $38$ {\tiny $\pm 5$}& $31$ {\tiny $\pm 23$} & $31$ {\tiny $\pm 23$}\\
\midrule
\texttt{cube-single-task1}  &$10$ {\tiny $\pm 5$}& $89$ {\tiny $\pm 5$}  & $97$ {\tiny $\pm 2$}& $97$ {\tiny $\pm 2$}& $96$ {\tiny $\pm 4$}& \mathbf{$99$} {\tiny $\pm 2$} & \mathbf{$99$} {\tiny $\pm 2$}\\
\texttt{cube-single-task2(*)}  &$3$ {\tiny $\pm 1$}& $92$ {\tiny $\pm 4$}  & \mathbf{$99$} {\tiny $\pm 0$}& $97$ {\tiny $\pm 2$}& $95$ {\tiny $\pm 3$}& \mathbf{$99$} {\tiny $\pm 2$} & \mathbf{$99$} {\tiny $\pm 2$}\\
\texttt{cube-single-task3}  &$9$ {\tiny $\pm 3$}& $93$ {\tiny $\pm 3$}  & \mathbf{$99$} {\tiny $\pm 1$}& $98$ {\tiny $\pm 2$}& $99$ {\tiny $\pm 1$}& $98$ {\tiny $\pm 3$} & $98$ {\tiny $\pm 3$}\\
\texttt{cube-single-task4}  &$2$ {\tiny $\pm 1$}& $92$ {\tiny $\pm 3$}  & $95$ {\tiny $\pm 2$}& $94$ {\tiny $\pm 3$}& $91$ {\tiny $\pm 10$}& \mathbf{$97$} {\tiny $\pm 3$} & \mathbf{$97$} {\tiny $\pm 3$}\\
\texttt{cube-single-task5}  &$3$ {\tiny $\pm 3$}& $87$ {\tiny $\pm 8$}  & $93$ {\tiny $\pm 3$}& $93$ {\tiny $\pm 3$}& $91$ {\tiny $\pm 5$}& \mathbf{$96$} {\tiny $\pm 4$} & \mathbf{$96$} {\tiny $\pm 4$}\\
\midrule
\texttt{cube-double-task1} \textcolor{pink}{\textit{(Hard)}}  &$8$ {\tiny $\pm 3$}& $45$ {\tiny $\pm 6$} & \mathbf{$77$} {\tiny $\pm 11$} & $61$ {\tiny $\pm 9$}& $63$ {\tiny $\pm 6$}& $50$ {\tiny $\pm 24$} & $50$ {\tiny $\pm 24$}\\
\texttt{cube-double-task2(*)} \textcolor{pink}{\textit{(Hard)}}  &$0$ {\tiny $\pm 0$}& $7$ {\tiny $\pm 3$} & $33$ {\tiny $\pm 8$} & $36$ {\tiny $\pm 6$}& $29$ {\tiny $\pm 8$}& \mathbf{$72$} {\tiny $\pm 15$} & \mathbf{$72$} {\tiny $\pm 15$}\\
\texttt{cube-double-task3} \textcolor{pink}{\textit{(Hard)}}  &$0$ {\tiny $\pm 0$}& $4$ {\tiny $\pm 1$} & $12$ {\tiny $\pm 6$} & $22$ {\tiny $\pm 5$}& $18$ {\tiny $\pm 6$}& \mathbf{$57$} {\tiny $\pm 14$} & \mathbf{$57$} {\tiny $\pm 14$}\\
\texttt{cube-double-task4} \textcolor{pink}{\textit{(Hard)}}  &$0$ {\tiny $\pm 0$}& $1$ {\tiny $\pm 1$} & $7$ {\tiny $\pm 4$} & $5$ {\tiny $\pm 2$}& $4$ {\tiny $\pm 2$}& \mathbf{$8$} {\tiny $\pm 4$} & \mathbf{$8$} {\tiny $\pm 4$}\\
\texttt{cube-double-task5} \textcolor{pink}{\textit{(Hard)}}  &$0$ {\tiny $\pm 0$}& $4$ {\tiny $\pm 2$} & $1$ {\tiny $\pm 1$} & $19$ {\tiny $\pm 10$}& $10$ {\tiny $\pm 5$}& \mathbf{$50$} {\tiny $\pm 11$} & \mathbf{$50$} {\tiny $\pm 11$}\\
\midrule
\texttt{scene-task1} &$19$ {\tiny $\pm 6$}& $95$ {\tiny $\pm 2$}  & $99$ {\tiny $\pm 1$} & \mathbf{$100$} {\tiny $\pm 0$}& $99$ {\tiny $\pm 1$}& \mathbf{$100$} {\tiny $\pm 1$} & \mathbf{$100$} {\tiny $\pm 1$}\\
\texttt{scene-task2(*)} &$1$ {\tiny $\pm 1$}& $50$ {\tiny $\pm 13$}  & \mathbf{$89$} {\tiny $\pm 9$} & $76$ {\tiny $\pm 9$}& $78$ {\tiny $\pm 7$}& $83$ {\tiny $\pm 10$} & $83$ {\tiny $\pm 10$}\\
\texttt{scene-task3} &$1$ {\tiny $\pm 1$}& $55$ {\tiny $\pm 16$}  & $97$ {\tiny $\pm 1$} & \mathbf{$98$} {\tiny $\pm 1$}& $96$ {\tiny $\pm 3$}& \mathbf{$98$} {\tiny $\pm 2$} & \mathbf{$98$} {\tiny $\pm 2$}\\
\texttt{scene-task4} &$2$ {\tiny $\pm 2$}& $3$ {\tiny $\pm 3$}  & $1$ {\tiny $\pm 1$} & $5$ {\tiny $\pm 1$}& \mathbf{$10$} {\tiny $\pm 6$}& $9$ {\tiny $\pm 7$} & $9$ {\tiny $\pm 7$}\\
\texttt{scene-task5} &$0$ {\tiny $\pm 0$}& $0$ {\tiny $\pm 0$}  & $0$ {\tiny $\pm 0$} & $0$ {\tiny $\pm 0$}& $0$ {\tiny $\pm 1$}& $0$ {\tiny $\pm 0$} & $0$ {\tiny $\pm 0$}\\
\midrule
\texttt{puzzle-3x3-task1} \textcolor{pink}{\textit{(Hard)}} &$5$ {\tiny $\pm 2$}& \mathbf{$97$} {\tiny $\pm 4$}& $-$ {\tiny $\pm -$}& $90$ {\tiny $\pm 4$}& $90$ {\tiny $\pm 6$}& $95$ {\tiny $\pm 4$} & $95$ {\tiny $\pm 4$}\\
\texttt{puzzle-3x3-task2} \textcolor{pink}{\textit{(Hard)}} &$1$ {\tiny $\pm 1$}& $1$ {\tiny $\pm 1$}& $-$ {\tiny $\pm -$}& $16$ {\tiny $\pm 5$}& $14$ {\tiny $\pm 2$}& \mathbf{$18$} {\tiny $\pm 10$} & \mathbf{$18$} {\tiny $\pm 10$}\\
\texttt{puzzle-3x3-task3} \textcolor{pink}{\textit{(Hard)}} &$1$ {\tiny $\pm 1$}& $3$ {\tiny $\pm 1$}& $-$ {\tiny $\pm -$}& $10$ {\tiny $\pm 3$}& $11$ {\tiny $\pm 3$}& \mathbf{$16$} {\tiny $\pm 7$} & \mathbf{$16$} {\tiny $\pm 7$}\\
\texttt{puzzle-3x3-task4(*)} \textcolor{pink}{\textit{(Hard)}} &$1$ {\tiny $\pm 1$}& $2$ {\tiny $\pm 1$}& $-$ {\tiny $\pm -$}& $16$ {\tiny $\pm 5$}& $14$ {\tiny $\pm 4$}& \mathbf{$17$} {\tiny $\pm 6$} & \mathbf{$17$} {\tiny $\pm 6$}\\
\texttt{puzzle-3x3-task5} \textcolor{pink}{\textit{(Hard)}} &$1$ {\tiny $\pm 0$}& $5$ {\tiny $\pm 3$}& $-$ {\tiny $\pm -$}& $16$ {\tiny $\pm 3$}& $16$ {\tiny $\pm 9$}& \mathbf{$38$} {\tiny $\pm 6$} & \mathbf{$38$} {\tiny $\pm 6$}\\
\midrule
\texttt{puzzle-4x4-task1} \textcolor{pink}{\textit{(Hard)}} &$1$ {\tiny $\pm 1$} & $26$ {\tiny $\pm 4$} & $-$ {\tiny $\pm -$}& $34$ {\tiny $\pm 8$}& $14$ {\tiny $\pm 4$}& $39$ {\tiny $\pm 7$ } & \mathbf{$47$} {\tiny $\pm 7$ }\\
\texttt{puzzle-4x4-task2} \textcolor{pink}{\textit{(Hard)}} &$0$ {\tiny $\pm 0$} & $12$ {\tiny $\pm 4$} & $-$ {\tiny $\pm -$}& $16$ {\tiny $\pm 5$}& $12$ {\tiny $\pm 3$}& $20$ {\tiny $\pm 4$ } & \mathbf{$21$} {\tiny $\pm 6$ }\\
\texttt{puzzle-4x4-task3} \textcolor{pink}{\textit{(Hard)}} &$0$ {\tiny $\pm 0$} & $15$ {\tiny $\pm 3$} & $-$ {\tiny $\pm -$}& $18$ {\tiny $\pm 5$}& $9$ {\tiny $\pm 4$}& $23$ {\tiny $\pm 6$ } & \mathbf{$36$} {\tiny $\pm 5$ }\\
\texttt{puzzle-4x4-task4(*)} \textcolor{pink}{\textit{(Hard)}} &$0$ {\tiny $\pm 0$} & $10$ {\tiny $\pm 3$} & $-$ {\tiny $\pm -$}& $11$ {\tiny $\pm 3$}& $5$ {\tiny $\pm 2$}& $12$ {\tiny $\pm 4$ } & \mathbf{$19$} {\tiny $\pm 5$ }\\
\texttt{puzzle-4x4-task5} \textcolor{pink}{\textit{(Hard)}} &$0$ {\tiny $\pm 0$} & $7$ {\tiny $\pm 3$} & $-$ {\tiny $\pm -$}& $7$ {\tiny $\pm 3$}& $5$ {\tiny $\pm 2$}& $12$ {\tiny $\pm 3$ } & \mathbf{$16$} {\tiny $\pm 7$ }\\
\midrule
\end{tabular}
\end{threeparttable}
}
\end{table*}